\makeatletter
\newsavebox{\arxivlayoutbox}
\newsavebox{\arxivauthorbox}
\AddToHook{begindocument/before}{%
  \iclrfinalcopy
  \author{%
    {\bfseries Yifei Zhao\textsuperscript{1,*,$\dagger$}\quad
    Fanyu Zhao\textsuperscript{1,*}\quad
    Han Zhang\textsuperscript{1}\quad
    Yixuan Lin\textsuperscript{1}}\\[0.5pt]
    {\bfseries Yuyang Hou\textsuperscript{1}\quad
    Bin Wu\textsuperscript{1}\quad
    Ruike Cao\textsuperscript{2}\quad
    Li Xiao\textsuperscript{2}\quad
    Yinsheng Li\textsuperscript{1,$\dagger$}}\\[1pt]
    {\fontsize{8}{9}\selectfont\textsuperscript{1}Fudan University\quad
    \textsuperscript{2}University of Science and Technology of China}\\[1pt]
    {\fontsize{8}{9}\selectfont\textsuperscript{*}Equal contribution.\quad
    \textsuperscript{$\dagger$}Corresponding authors.\quad
    \href{mailto:yfzhao19@fudan.edu.cn}{\texttt{yfzhao19@fudan.edu.cn}}}%
  }%
  \hypersetup{%
    hypertexnames=false,
    pdftitle={Do 3D Large Language Models Really Use 3D Evidence?},
    pdfauthor={Yifei Zhao, Fanyu Zhao, Han Zhang, Yixuan Lin, Yuyang Hou, Bin Wu, Ruike Cao, Li Xiao, Yinsheng Li}%
  }%
  \fancyhead{}%
  \renewcommand{\headrulewidth}{0pt}%
  \renewcommand{\@maketitle}{%
    \vbox{\hsize\textwidth
      {\LARGE\sc\@title\par}%
      \sbox{\arxivlayoutbox}{%
        \begin{tabular}[t]{l}\bf\rule{\z@}{24pt}%
          Anonymous authors\\Paper under double-blind review
        \end{tabular}}%
      \sbox{\arxivauthorbox}{%
        \begin{minipage}[b]{\textwidth}%
          \centering\fontsize{9}{10}\selectfont
          \setlength{\parskip}{0pt}%
          \@author
        \end{minipage}}%
      \raisebox{\dimexpr\dp\arxivauthorbox-\dp\arxivlayoutbox-8pt\relax}%
        [\ht\arxivlayoutbox][\dp\arxivlayoutbox]{\usebox{\arxivauthorbox}}%
      \vskip 0.3in minus 0.1in
    }%
  }%
}
\makeatother

\documentclass{article}
\usepackage{iclr2027_conference,times}

\usepackage{amsmath,amsfonts,bm}

\def\eqref#1{equation~\ref{#1}}

\def\1{\bm{1}}

\DeclareMathAlphabet{\mathsfit}{\encodingdefault}{\sfdefault}{m}{sl}
\SetMathAlphabet{\mathsfit}{bold}{\encodingdefault}{\sfdefault}{bx}{n}

\usepackage{hyperref}
\usepackage{url}
\usepackage{booktabs}
\usepackage{graphicx}
\usepackage{float}
\usepackage{placeins}
\usepackage{needspace}
\usepackage{wrapfig}
\usepackage{amsmath}
\usepackage{amssymb}
\usepackage{multirow}
\usepackage{xcolor}
\usepackage{colortbl}
\usepackage{array}
\usepackage{makecell}
\usepackage{tabularx}
\usepackage{pifont}
\usepackage[percent]{overpic}
\usepackage{tikz}
\usetikzlibrary{shapes.geometric}
\usepackage{pgfplots}
\pgfplotsset{compat=1.17}

\title{Do 3D Large Language Models Really Use 3D Evidence?}

\definecolor{farred}{RGB}{173,45,45}
\definecolor{fargreen}{RGB}{37,120,83}
\definecolor{farblue}{RGB}{45,83,150}
\definecolor{mainrankfirst}{RGB}{120,190,142}
\definecolor{mainranksecond}{RGB}{179,218,190}
\definecolor{mainrankthird}{RGB}{225,240,229}
\newcommand{\rankfirst}[1]{\cellcolor{mainrankfirst}#1}
\newcommand{\ranksecond}[1]{\cellcolor{mainranksecond}#1}
\newcommand{\rankthird}[1]{\cellcolor{mainrankthird}#1}
\hypersetup{colorlinks=true,citecolor=farblue,linkcolor=farblue,urlcolor=farblue}
\newcommand{\far}{\textsc{Unicorns}}
\newcommand{\horn}{\textsc{Horn}}
\newsavebox{\hornablationbox}
\newcommand{\good}[1]{\textcolor{fargreen}{#1}}
\newcommand{\bad}[1]{\textcolor{farred}{#1}}

\newcommand{\backtomain}[1]{%
  \par\begingroup\normalfont\normalsize\bfseries\centering
  \setlength{\parskip}{0pt}\vspace{0.5\baselineskip}
  \noindent\hyperref[#1]{Back to Section~\ref*{#1}: \nameref*{#1}}\par
  \endgroup}
\newcommand{\ms}[2]{#1{\tiny\,$\pm$\,#2}}

\begin{document}
\maketitle

\begin{abstract}
Recent 3D Large Language Models (3D-LLMs) have shown strong performance on scene question answering with extensive post-training.
Yet, existing benchmarks rarely evaluate how 3D-LLMs respond to \textit{referential ambiguity, insufficient scene evidence, or false presuppositions}, obscuring whether their responses are grounded in 3D evidence.
In this work, we introduce UNICORNS, a benchmark for evaluating the reliability of 3D-LLM responses across four evidence conditions.
UNICORNS contains 11,387 questions across 557 ScanNet scans, covering eight challenge categories alongside answerable controls.
A response passes only if it satisfies the required response behavior and contains no scene claim unsupported by the available evidence.
Evaluation of six open-sourced 3D-LLMs reveals a pronounced performance gap: even the strongest zero-shot model achieves 61.22\% accuracy on answerable controls, but only 40.94\% across the eight challenge categories.
To address this gap, we further propose \horn{}, an obligation-aware post-training method combining group-relative task rewards, leaf-conditioned obligation rewards, and supervised control replay.
Extensive experiments demonstrate that \horn{} reaches $75.1{\pm}1.0\%$ across the challenge categories and $77.2{\pm}0.8\%$ overall, outperforming all baselines without degrading answerable-control accuracy.
Our findings enable a systematic evaluation and improvement of evidence-grounded 3D reasoning for 3D-LLMs (\href{https://anonymous.4open.science/r/unicorns-6B38/}{code}).
\end{abstract}

\section{Introduction}

3D large language models (3D-LLMs) extend pretrained language models to 3D scene understanding and reasoning by integrating representations such as point clouds, object-level features, and multiview observations~\citep{hong2023ll3da,huang2024leo,huang2024chatscene}.
These models support applications including 3D dialogue, task planning, and embodied action~\citep{huang2024leo}.
Reliable 3D reasoning, however, requires models not only to answer questions but also to recognize what the available observations do and do not support~\citep{wu2026abstaineqa}.
In realistic settings, questions may contain ambiguous references, ask about unobserved information, or rest on premises contradicted by scene evidence~\citep{wu2026abstaineqa}.
Models may nevertheless produce fluent, plausible responses whose claims are unsupported by the scene.
Existing 3D QA benchmarks predominantly focus on questions with determinate answers and rarely evaluate models under these conditions; moreover, when scene evidence and language priors favor the same response, correctness alone cannot distinguish evidence-grounded reasoning from reliance on language priors~\citep{ma2026real3dqa}.

\begin{figure}
\centering
\includegraphics[width=\textwidth]{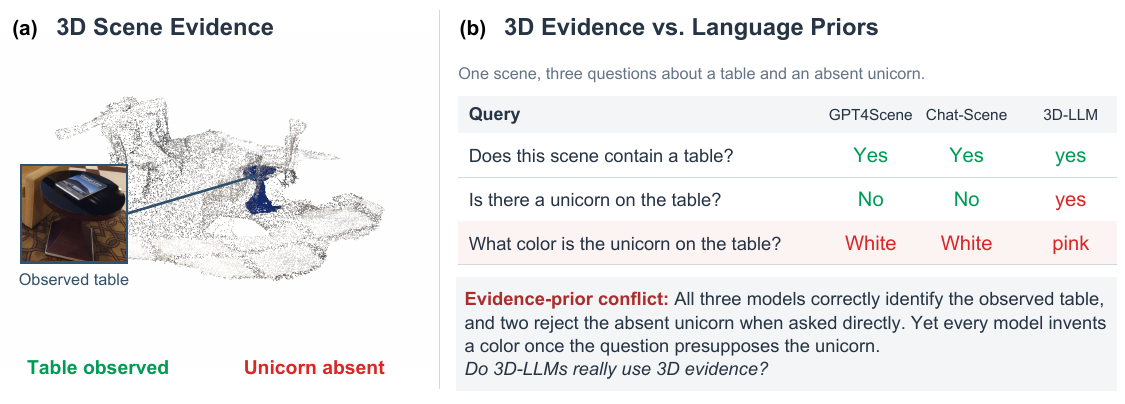}
\vspace{-4mm}
\caption{All models recognize the table (highlighted in blue). Two of them reject the absent unicorn, yet all invent its color under a presupposing question.}
\label{fig:unicorn}
\end{figure}

Figure~\ref{fig:unicorn} illustrates this limitation.
Given the same scene, all three models recognize the table, and two correctly deny that a unicorn is present when asked directly.
Yet, when the absent object is embedded in a presupposing attribute question, all three assign it a color.
This sensitivity to question framing despite unchanged scene evidence motivates our central question: \emph{do 3D-LLMs really use 3D evidence?}
Prior work addresses complementary aspects of this problem.
SceneCOT grounds intermediate reasoning in scene objects~\citep{linghu2026scenecot}. Real-3DQA filters text-answerable questions and tests viewpoint-rotation consistency~\citep{ma2026real3dqa}. AbstainEQA benchmarks image-based abstention under referential underspecification, unavailable information, and false presuppositions~\citep{wu2026abstaineqa}.
However, none jointly evaluates whether response strategies fit the evidence and all response claims are supported, even an appropriate strategy may introduce unsupported details.
We therefore assess both dimensions, with answerable controls for sufficient-evidence questions.

To fill this gap, we introduce \far{}, a benchmark that jointly evaluates evidence-appropriate response behavior and factual support for complete responses. The benchmark contains 11,387 questions spanning 557 ScanNet scans~\citep{dai2017scannet} and 353 object classes. Eight challenge categories and their paired answerable controls cover four evidence states. Scene-disjoint splits and executable inventory and geometry checks establish the evidence available for each question. Three LLM judges~\citep{openaigpt5,googledeepmind2026gemini31pro,deepseekai2024deepseekv3} apply a shared evidence-veto policy by majority vote, passing a response only if it follows the required behavior and makes no unsupported scene claims.

We further propose \horn{} (\textbf{H}eterogeneous \textbf{O}bligation \textbf{R}ewards with \textbf{N}on-refusal replay), an obligation-aware post-training method combining group-relative task rewards, leaf-conditioned obligation rewards, and supervised control replay.
Across six released 3D-LLMs, the strongest zero-shot model achieves 61.22\% accuracy on answerable controls but only 40.94\% across the challenge categories.
Starting from the same SceneCOT SFT checkpoint, \horn{} reaches $75.1{\pm}1.0\%$ Challenge, $79.3{\pm}0.7\%$ Control, and $77.2{\pm}0.8\%$ Overall, outperforming matched GRPO by 5.1 and 3.2 percentage points on Challenge and Overall, respectively.
Controlled scene edits, Ego-Visible evaluation, and text-only tests further assess evidence sensitivity, robustness to partial observations, and dependence on 3D input.
Our main contributions are summarized as:
\begin{enumerate}
\item \textbf{The \far{} benchmark.} A point-cloud-native 3D benchmark that jointly evaluates evidence-appropriate response behavior and complete-response factual support across four evidence states, with executable scene checks and answerable controls.
\item \textbf{Systematic evaluation and analysis.} An evaluation of six released 3D-LLMs with response-strategy and factual-support diagnostics, controlled scene edits, Ego-Visible testing, and text-only ablations.
\item \textbf{The \horn{} post-training method.} An obligation-aware method that outperforms matched SFT and GRPO on Challenge and Overall accuracy while preserving performance on answerable controls.
\end{enumerate}

\section{Related Work}\label{sec:related}

\textbf{Grounded 3D-LLMs reasoning.}
3D-LLM connects 3D features with language models, while LEO and Chat-Scene incorporate object features and multiview observations to support scene understanding and embodied tasks~\citep{hong2023ll3da,huang2024leo,huang2024chatscene}. SceneCOT further grounds intermediate reasoning in specific objects and checks consistency between grounding and answers~\citep{linghu2026scenecot}. Real-3DQA's reweighted fine-tuning targets language shortcuts by emphasizing predictions that depend on 3D input~\citep{ma2026real3dqa}. These methods improve scene representations and reasoning, motivating evaluation of whether their responses remain reliable under different evidence conditions.

\textbf{Ambiguity, insufficient evidence, and response strategies.}
Visual questions may require clarification or abstention when the available evidence does not determine an answer.
ClearVQA trains interactive clarification~\citep{jian2025clearvqa}, AQuA distinguishes direct answering, inference, enumeration, and clarification~\citep{jang2026aqua}, and CoA learns when to answer or ask~\citep{cao2026coa}.
AbstainEQA studies abstention under missing information, referential underspecification, and false presuppositions, with paired answerable questions, on egocentric RGB frames rather than 3D point clouds~\citep{wu2026abstaineqa}.
ProactiveBench and Ask-to-Act extend this perspective to requesting evidence and clarifying instructions before action~\citep{demin2026proactivebench,ramrakhya2025asktoact}.
Together, these studies motivate adapting response strategies to evidence conditions while preserving ordinary answering.

\textbf{Diagnostic benchmarks for response reliability.}
Existing 3D benchmarks assess different aspects of scene understanding. ScanQA, SQA3D, and MSR3D focus on scene question answering and situated reasoning~\citep{azuma2022scanqa,ma2023sqa3d,linghu2024msr3d}. Real-3DQA probes language shortcuts and viewpoint consistency~\citep{ma2026real3dqa}, and Ambi3D evaluates instruction ambiguity detection~\citep{ding2026ambi3d}.
These benchmarks do not jointly assess response strategies across evidence conditions and factual support for complete responses.
\far{} unifies both requirements into a single pass criterion that rejects any unsupported scene claim, even when the strategy is appropriate.
Table~\ref{tab:prior-protocols} summarizes these differences in evaluation coverage.

\begingroup
\begin{table}
\centering
\fontsize{8}{9}\selectfont
\setlength{\tabcolsep}{3pt}
\renewcommand{\arraystretch}{1.15}
\newcommand{\benchyes}{\good{\ding{51}}}
\newcommand{\benchno}{\bad{\ding{55}}}
\caption{\textbf{Comparison of representative benchmarks with 3D input.} Claim-level veto fails a  response if any scene claim lacks support.}
\label{tab:prior-protocols}
\begin{tabularx}{\textwidth}{@{}l>{\raggedright\arraybackslash}Xccccc@{}}
\toprule
Benchmark & Evaluation target & Ambiguity & \makecell{Unobserved\\evidence} & \makecell{False\\premise} & \makecell{Answerable or\\unambiguous} & \makecell{Claim-level\\veto} \\
\midrule
ScanQA & Scene QA & \benchno & \benchno & \benchno & \benchyes & \benchno \\
SQA3D & Situated QA & \benchno & \benchno & \benchno & \benchyes & \benchno \\
Real-3DQA & Shortcuts; viewpoint consistency & \benchno & \benchno & \benchno & \benchyes & \benchno \\
Ambi3D & Binary ambiguity detection & \benchyes & \benchno & \benchno & \benchyes & \benchno \\
\rowcolor{farblue!7}
\textbf{\far{}} & \textbf{Complete-response reliability} & \benchyes & \benchyes & \benchyes & \benchyes & \benchyes \\
\bottomrule
\end{tabularx}
\end{table}
\endgroup

\section{The \texorpdfstring{\far{}}{UNICORNS} Benchmark}
\label{sec:benchmark}
This study proposes \far{} to jointly assess response strategies and the factual reliability of complete responses in 3D scenes.
Models are required to respond according to the available evidence and ground every scene claim in it.
Full Scene is the primary evaluation setting, using the complete scene point cloud.

\subsection{Evidence States and Challenge Taxonomy}\label{sec:taxonomy}
We define four states by the relationship between a question and scene evidence: determinate, ambiguous, unobserved, and contradicted.
They respectively require direct answering, clarification or candidate enumeration, an explicit account of what cannot be determined, and rejection or correction of a false premise.
Eight challenge leaves cover ambiguity (U/N), unobserved information (I/C), and contradiction (O/R/N$^\prime$/S). Table~\ref{tab:taxonomy} gives their definitions and required responses.
Their initials spell \far{}, with the prime distinguishing the second N.
Answerable Control samples serve as the basis for constructing paired Challenges, testing whether models retain correct direct answering alongside challenge handling.

\begin{table}
\centering
\small
\caption{Challenge categories, answerable controls, and required response behaviors.}
\label{tab:taxonomy}
\begin{tabularx}{\linewidth}{@{}c@{\hspace{4pt}}l@{\hspace{6pt}}X@{}}
\toprule
Letter & Category & Required behavior \\
\midrule
\textbf{U} & Underspecified class-only reference & Identify multiple same-class candidates; clarify or enumerate. \\
\textbf{N} & Non-disambiguating modifier & Explain modifier ambiguity; clarify or enumerate. \\
\textbf{I} & Information beyond scan coverage & State that the scan cannot determine the answer. \\
\textbf{C} & Closed-container interior & Do not assert unseen contents; express bounded uncertainty. \\
\textbf{O} & Object-class absence & Reject or correct the existence presupposition. \\
\textbf{R} & Refuted attribute claim & Reject or correct the attribute without inventing a replacement. \\
\textbf{\(\mathrm{N}^{\prime}\)} & Non-holding spatial relation & Reject/correct the relation using verified geometry. \\
\textbf{S} & Size-incompatible placement & Reject the action; explain the geometric conflict if useful. \\
-- & Answerable control & Give the direct scene-grounded answer; refusal fails. \\
\bottomrule
\end{tabularx}
\end{table}

\subsection{Data Construction and Splits}\label{sec:dataset}
We construct questions, supporting evidence, and response requirements from ScanNet annotations and geometry~\citep{dai2017scannet}.
Each leaf draws on prior construction principles without copying released QA samples. \hypertarget{main:app:lineage}{}\label{main:app:lineage}Appendix~\ref{app:lineage} maps these principles to their sources.
Independent verification programs recompute candidate cardinality, existence, attributes, relations, visibility, and fit ratios (\hypertarget{main:app:contracts}{}\label{main:app:contracts}Appendix~\ref{app:contracts}).
Samples with incomplete evidence, conflicting sources, or same-scene semantic or normalized-question duplicates are excluded.
Each sample retains scene/source identifiers, its construction program, evidence, response obligation, and rubric for traceability.
The scene-disjoint split contains 8,251 training instances, including 3,300 Control samples, and 3,136 test instances.
The test set comprises 1,568 Challenge samples (196 per leaf) and 1,568 Control samples.
\hypertarget{main:app:splits}{}\label{main:app:splits}Appendix~\ref{app:splits} gives scene totals and training distributions.

\begin{figure}[H]
\centering
\includegraphics[width=\linewidth]{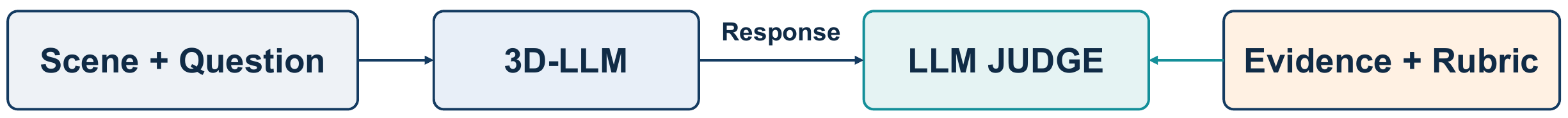}
\caption{\textbf{\far{} evaluation pipeline.} The LLM judge assesses both response strategy and factual support using benchmark evidence and the response rubric.}
\label{fig:benchmark-pipeline}
\end{figure}

\subsection{Evaluation Protocol}\label{sec:metrics}
Figure~\ref{fig:benchmark-pipeline} summarizes our evaluation pipeline.
A 3D-LLM receives scene observations and a question, while an ensemble of three LLM judges evaluates its response using the question, structured scene evidence, and a response rubric.

Each instance is represented as $x=(S,q,e,o,r,z,\ell)$: scene $S$, question $q$, machine-checkable evidence $e$, response obligation $o$, response rubric $r$, evidence state $z$, and challenge leaf or Control $\ell$.
Models generate free-form text $y$ from the scene and question.
We define two Boolean predicates: $\mathrm{Strategy}(y,o,r)$ checks whether the response follows the required behavior, and $\mathrm{Factual}(y,e,r)$ checks both the required factual content and the support for every scene claim.
A response passes only if both hold:
\begin{equation}
\mathrm{Adm}(y\mid x)=\mathrm{Strategy}(y,o,r)\land\mathrm{Factual}(y,e,r).
\end{equation}
Three LLM judges independently apply the same leaf-aware prompt (Appendix~\ref{app:judge-prompts}) using the question, response, leaf, obligation, rubric, and structured evidence.
Each judge returns strategy and factual verdicts; a response passes if a majority of judges pass both on that response.
Different valid formulations can pass, but any unsupported scene claim vetoes the complete response, even with an appropriate strategy.
For example, uncertainty followed by an unsupported count, color, or relation fails.
Controls require a correct direct answer, so refusal fails.

\textbf{Evaluation metrics.}
For any evaluation subset $D$, Strategic, Factual, and Veto Accuracy are defined as the mean values of $\mathrm{Strategy}$, $\mathrm{Factual}$, and $\mathrm{Adm}$, respectively. Unless otherwise stated, accuracy refers to Veto Accuracy. Let $A_\ell$ and $A_{\mathrm{C0}}$ denote Veto Accuracy on challenge leaf $\ell$ and Controls, respectively. We report per-leaf accuracy together with $\mathrm{Challenge}=\frac{1}{8}\sum_{\ell\in\mathcal L}A_\ell$, $\mathrm{Control}=A_{\mathrm{C0}}$, and $\mathrm{Overall}=\frac{1}{2}(\mathrm{Challenge}+\mathrm{Control})$; the latter equality follows from the balanced test design.

\section{\texorpdfstring{\horn{}}{HORN}: Obligation-Aware Post-Training}\label{sec:method}
\begin{figure}
\centering
\includegraphics[width=\linewidth]{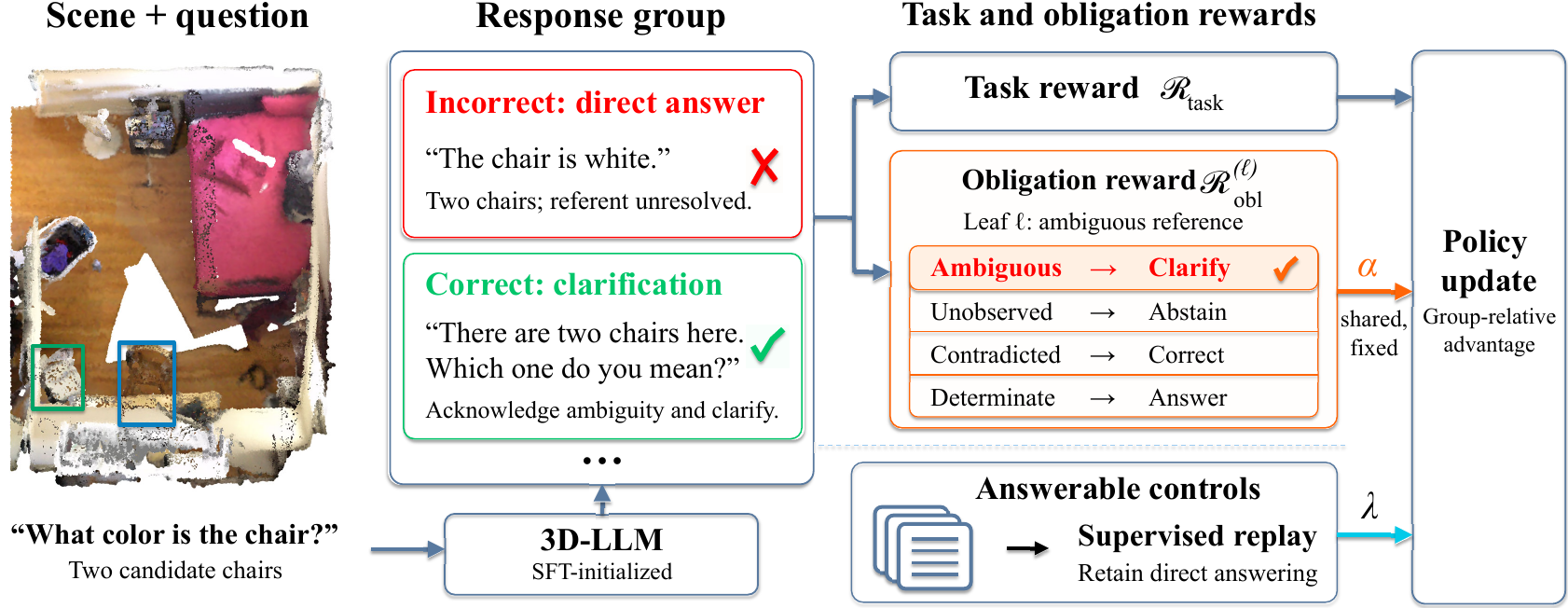}
\caption{\textbf{The overview of \horn{}.} An SFT-initialized 3D-LLM generates response groups. Task and leaf-conditioned obligation rewards guide group-relative policy updates with a shared, fixed coefficient $\alpha$. Supervised control replay, weighted by $\lambda$, supports direct answering.}
\label{fig:horn-method}
\end{figure}

To improve the response reliability examined by \far{}, we introduce \horn{}, which combines task rewards, leaf-conditioned obligation rewards, and supervised control replay on an SFT-initialized 3D-LLM (Figure~\ref{fig:horn-method}).
It learns when to clarify, express uncertainty, or correct a premise while retaining direct answering when evidence is sufficient (Appendix~\ref{app:horn-reproduction}).

\textbf{Reward construction.}\label{sec:horn-rewards}\label{sec:horn-weighting}
For each scene--question pair $(S,q)$ in instance $x$, the policy $\pi_\theta(y\mid S,q)$ generates $G=6$ responses.
The task reward checks response strategy and factual content against reference answers and structured evidence, and computes grounding and format scores $g_i$ and $f_i$.
These checks use rule-based scoring, with response length incorporated into the format score.
Task scores are assigned automatically from within-group comparisons.
Responses are ordered by three verification levels: strategy failure, strategy-only success, and joint strategy and factual success.
Strategy failures are tied at the bottom; within the other levels, grounding is compared first and format second, with ties assigned average ranks.
Let $u_i\in\{0,1\}$ indicate strategy success and let ranks increase from worst to best, from $1$ to $G$.
The task reward is
\begin{equation}
R_{\mathrm{task}}(x,y_i)=u_i\frac{\operatorname{rank}(y_i)-1}{G-1}.
\label{eq:horn-rewards}
\end{equation}
This construction assigns rewards through within-group comparisons, while giving strategy failures zero task reward.
The obligation reward $R_{\mathrm{obl}}^{(\ell)}$ provides a complementary signal for the behavior required by leaf $\ell$: clarification or enumeration for ambiguity, bounded uncertainty for unobserved information, and rejection or correction for contradicted premises.
This reward uses leaf-specific rules that account for response length.
Both coordinates provide training signals; the separate LLM judge evaluates the complete response.

\textbf{Policy optimization and control replay.}\label{sec:horn-replay}
We center each reward coordinate within its response group and combine them using a shared, fixed global obligation coefficient $\alpha$ to obtain advantages for group-relative policy updates~\citep{shao2024deepseekmath}.
Groups with no variation in either coordinate are resampled and skipped if they remain uninformative after the resampling limit.
On effective challenge updates, we also draw an answerable-control batch $B_{\mathrm{C0}}$ from the training pool and add supervised cross-entropy to provide continued direct-answer supervision.
The objective is
\begin{equation}
\mathcal L_{\mathrm{HORN}}=\mathcal L_{\mathrm{clipped}}+\beta_{\mathrm{KL}}\mathcal L_{\mathrm{ref}}+\lambda\,\mathbf 1[\ell\ne\mathrm{C0}]\mathcal L_{\mathrm{SFT}}(\theta;B_{\mathrm{C0}}).
\label{eq:horn-objective}
\end{equation}
Here $\mathcal L_{\mathrm{clipped}}$ is the clipped policy loss, $\mathcal L_{\mathrm{ref}}$ is the reference-policy regularizer, $\beta_{\mathrm{KL}}$ controls regularization strength, and $\lambda$ controls supervised replay; replay is omitted when no control batch is available.
The model retains its native scene representation.
Leaf labels, references, and structured evidence provide training supervision only; inference requires the scene and question.

\section{Experiments}
\subsection{Experimental Setup}\label{sec:experimental-setup}
\textbf{Baselines.}
We evaluate six released 3D-LLMs: LEO~\citep{huang2024leo}, Chat-Scene~\citep{huang2024chatscene}, Chat-3D-v2~\citep{huang2023chat3dv2}, 3D-LLM~\citep{hong2023ll3da}, GPT4Scene~\citep{qi2025gpt4scene}, and SceneCOT~\citep{linghu2026scenecot}, with their native scene representations. \textbf{Zero-shot} rows use released checkpoints without \far{} post-training. \textbf{Supervised fine-tuning (SFT).} All 20\% configurations use the same fixed 1,650-example subset, while full-data SceneCOT SFT serves only as a data-scaling reference. We reimplement 3D-RFT~\citep{ma2026real3dqa} on SceneCOT because its complete official training pipeline is unavailable. \textbf{Policy optimization} defaults to 20\% SFT. Standard GRPO with a scalar task reward~\citep{shao2024deepseekmath} initializes GPT4Scene and SceneCOT from their respective 20\% SFT checkpoints and samples the full training pool.

\textbf{Evaluation metrics.}
Following Section~\ref{sec:metrics}, our primary measure is Veto Accuracy: a response passes if a majority of judges pass both strategy and factual support on that response. All three judges agree on 96.7\% of responses (Appendix~\ref{sec:judge-validation}).
We report per-leaf accuracy and the balanced Challenge, Control, and Overall aggregates.

\textbf{Implementation details.}
Starting from the SceneCOT 20\% SFT checkpoint, \horn{} runs 80 policy updates with six responses per question, using $\alpha=0.25$, $\lambda=0.10$, and $\beta_{\mathrm{KL}}=0.06$. Post-trained results are means $\pm$ sample SDs over three seeds; zero-shot uses one run. Fixed judgment criteria apply to the same 3,136 ID-aligned test instances for every method. \hypertarget{main:sec:judge-validation}{}\label{main:sec:judge-validation}Appendix~\ref{sec:judge-validation} specifies the judges and a blind human audit. Appendix~\ref{app:compute} reports hardware and training time.

\subsection{Main Results}\label{sec:main-results}
\begin{table*}
\centering
\footnotesize
\setlength{\tabcolsep}{2.0pt}
\caption{Veto Accuracy (\%) on \far{}. Post-training results are three-seed means; aggregates show mean $\pm$ sample SD. Green marks each column's top three: \colorbox{mainrankfirst}{1st}, \colorbox{mainranksecond}{2nd}, \colorbox{mainrankthird}{3rd}.}
\label{tab:main}
\setlength{\aboverulesep}{0pt}
\setlength{\belowrulesep}{0pt}
\renewcommand{\arraystretch}{1.08}
\begin{tabular*}{\textwidth}{@{\extracolsep{\fill}}lrrrrrrrr@{\hspace{4pt}}rr@{\hspace{4pt}}r@{}}
\toprule
& \multicolumn{8}{c}{\textbf{Challenge leaves}} & \multicolumn{3}{c}{\textbf{Aggregate metrics}} \\
\cmidrule(lr){2-9}\cmidrule(l){10-12}
\textbf{Model} & \textbf{U} & \textbf{N} & \textbf{I} & \textbf{C} & \textbf{O} & \textbf{R} & \(\mathrm{N}^{\prime}\) & \textbf{S} & \textbf{Challenge} & \textbf{Control} & \textbf{Overall} \\
\midrule
\rowcolor[gray]{.9}\multicolumn{12}{@{}l}{\emph{\textbf{Zero-shot}}} \\[-1pt]
LEO & 0.00 & 0.00 & 2.04 & 5.10 & \ranksecond{82.65} & 4.08 & 0.00 & 3.57 & 12.18 & 32.33 & 22.26 \\
3D-LLM & 0.51 & 1.53 & 4.59 & 2.04 & 3.06 & 20.92 & 7.65 & 59.69 & 12.50 & 47.13 & 29.82 \\
Chat-3D-v2 & 3.06 & 3.06 & 0.51 & 9.69 & 53.57 & 4.08 & 1.53 & 34.18 & 13.71 & 53.25 & 33.48 \\
Chat-Scene & 9.18 & 10.20 & 2.55 & 17.86 & 66.84 & 3.57 & 1.02 & 29.08 & 17.54 & 51.47 & 34.50 \\
SceneCOT & 7.14 & 20.41 & 19.39 & 35.20 & 70.41 & 52.04 & 8.67 & \rankfirst{99.49} & 39.09 & 30.80 & 34.95 \\
GPT4Scene & 8.67 & 7.65 & 15.31 & 42.35 & 79.59 & \rankfirst{69.90} & 18.88 & 85.20 & 40.94 & 61.22 & 51.08 \\
\addlinespace[2pt]
\rowcolor[gray]{.9}\multicolumn{12}{@{}l}{\emph{\textbf{Supervised fine-tuning}}} \\[-1pt]
GPT4Scene (20\%) & 11.22 & 14.29 & 14.80 & 41.84 & 81.12 & \rankthird{60.20} & 37.24 & 91.33 & \ms{44.0}{0.7} & \ms{74.3}{0.8} & \ms{59.1}{0.8} \\
3D-RFT (20\%) & 48.47 & 54.59 & \rankthird{89.80} & 47.45 & 69.90 & 45.41 & \ranksecond{97.96} & 93.37 & \ms{68.4}{0.8} & \ms{71.2}{0.9} & \ms{69.8}{0.9} \\
SceneCOT (100\%) & 26.53 & 25.00 & 86.22 & \rankfirst{85.20} & 73.47 & 36.22 & \rankthird{96.94} & 95.41 & \ms{65.6}{0.8} & \ms{77.2}{0.8} & \ms{71.4}{0.7} \\
SceneCOT (20\%) & \rankthird{50.51} & \rankthird{59.18} & 89.29 & 46.43 & 64.80 & 44.90 & 96.43 & \rankthird{96.94} & \rankthird{\ms{68.6}{0.8}} & \ranksecond{\ms{79.2}{0.6}} & \rankthird{\ms{73.9}{0.7}} \\
\addlinespace[2pt]
\rowcolor[gray]{.9}\multicolumn{12}{@{}l}{\emph{\textbf{Policy optimization}}} \\[-1pt]
GPT4Scene + GRPO & 8.67 & 14.29 & 15.31 & 42.35 & \rankfirst{83.16} & \ranksecond{62.76} & 28.06 & 91.33 & \ms{43.2}{0.8} & \ms{76.0}{0.9} & \ms{59.6}{0.8} \\
GPT4Scene + \horn{} & 11.22 & 15.48 & 14.80 & 39.29 & \rankthird{81.63} & 59.69 & 39.29 & 92.35 & \ms{44.2}{0.7} & \ms{76.8}{0.9} & \ms{60.5}{0.8} \\
SceneCOT + GRPO & \ranksecond{56.12} & \ranksecond{60.20} & \ranksecond{90.31} & \rankthird{50.00} & 64.29 & 46.94 & 94.90 & \rankthird{96.94} & \ranksecond{\ms{70.0}{1.0}} & \rankthird{\ms{78.1}{0.8}} & \ranksecond{\ms{74.0}{0.9}} \\
\textbf{\horn{} (Ours)} & \rankfirst{62.24} & \rankfirst{65.82} & \rankfirst{95.41} & \ranksecond{58.67} & 68.37 & 52.55 & \rankfirst{98.98} & \ranksecond{98.47} & \rankfirst{\ms{75.1}{1.0}} & \rankfirst{\ms{79.3}{0.7}} & \rankfirst{\ms{77.2}{0.8}} \\
\bottomrule
\end{tabular*}
\end{table*}

\textbf{Limitations of existing 3D-LLMs.}
Table~\ref{tab:main} reports zero-shot(ZS) Challenge accuracies of 12.18--40.94\%.
The strongest model, GPT4Scene, reaches 61.22\% on Control but only 40.94\% on Challenge, a gap of 20.28 percentage points (pp).
Thus, \emph{strong performance on ordinary answerable questions can give the appearance that models use 3D evidence}, yet their challenge responses are often unfaithful to that evidence.
Zero-shot SceneCOT reaches 39.09\% Challenge but only 30.80\% Control, underscoring the need to evaluate challenge handling and ordinary answering separately.
These results reveal substantial limitations in evidence-grounded response reliability across evidence conditions.
We therefore select the two strongest zero-shot models, GPT4Scene and SceneCOT, for in-depth analysis.
\hypertarget{main:app:split-gap}{}\label{main:app:split-gap}Appendix~\ref{app:split-gap} visualizes Challenge--Control performance gaps across models.

\textbf{Effectiveness of supervised fine-tuning.}
Table~\ref{tab:main} shows clear SFT gains on \far{}.
SceneCOT gains $+29.5$/$+48.4$~pp on Challenge/Control, and GPT4Scene $+3.1$/$+13.1$~pp.
These gains show that \far{} can improve the model's robustness to ambiguous and false-premise questions.
Figure~\ref{fig:posttrain-gain} shows category-wise SFT gains.

\textbf{Policy optimization and \horn{}.}
However, Figure~\ref{fig:posttrain-gain} shows that although standard GRPO~\citep{shao2024deepseekmath} raises SceneCOT's Challenge from 68.6\% to 70.0\%, Control drops to 78.1\%, and the R/S losses after SFT remain; GPT4Scene's Challenge falls from 44.0\% to 43.2\%.
Thus GRPO's gains vary by model and capability, and a scalar task reward does not restore the failed leaves.
With leaf-conditioned obligation rewards, resampling, and control replay (Section~\ref{sec:method}), \textbf{\horn{}} outperforms GRPO on both backbones: Challenge/Control/Overall improve by $+5.1$/$+1.2$/$+3.2$~pp on SceneCOT and $+1.0$/$+0.8$/$+0.9$~pp on GPT4Scene (Table~\ref{tab:main}).
It is the only SceneCOT post-training run that recovers R relative to zero-shot and nearly closes the S drop, with the lowest across-leaf variance (\hypertarget{main:app:category-balance}{}\label{main:app:category-balance}Appendix~\ref{app:category-balance}).
In Table~\ref{tab:horn-ablation}, removing $R_{\mathrm{obl}}$ lowers Challenge from 75.1\% to 70.6\% with replay and resampling retained, supporting the added scoring signal in this setting.
Without replay, Challenge peaks at 77.1\% but Control collapses to 68.5\%, as obligation-driven updates overfit refusal and forget ordinary answering.
Without resampling, both splits stay close to full \horn{} (74.7\%/78.8\% vs.\ 75.1\%/79.3\%), yet groups with no reward contrast can collapse a leaf's training signal.
Only the full combination raises Challenge by 6.5~pp over SFT while matching Control (79.3\% vs.\ 79.2\%); \hypertarget{main:app:compute}{}\label{main:app:compute}Appendix~\ref{app:compute} compares compute.

\sbox{\hornablationbox}{%
\setlength{\tabcolsep}{1.5pt}%
\renewcommand{\arraystretch}{1.263}%
\small
\begin{tabular}{@{}lccc@{}}
\toprule
Method & Challenge & Control & Overall \\
\midrule
SFT (20\%) & 68.6 & 79.2 & 73.9 \\
GRPO & 70.0 & 78.1 & 74.0 \\
\midrule
w/o $R_{\mathrm{obl}}$ & 70.6 & 78.9 & 74.8 \\
w/o replay & 77.1 & 68.5 & 72.8 \\
w/o resample & 74.7 & 78.8 & 76.8 \\
\textbf{\horn{}} & \textbf{75.1} & \textbf{79.3} & \textbf{77.2} \\
\bottomrule
\end{tabular}}

\begin{figure*}[t]
\centering
\setlength{\abovecaptionskip}{3pt}
\setlength{\belowcaptionskip}{0pt}
\begin{minipage}[t]{0.57\textwidth}
\vspace{0pt}
\centering
\includegraphics[width=\linewidth,trim=0 12bp 0 0,clip]{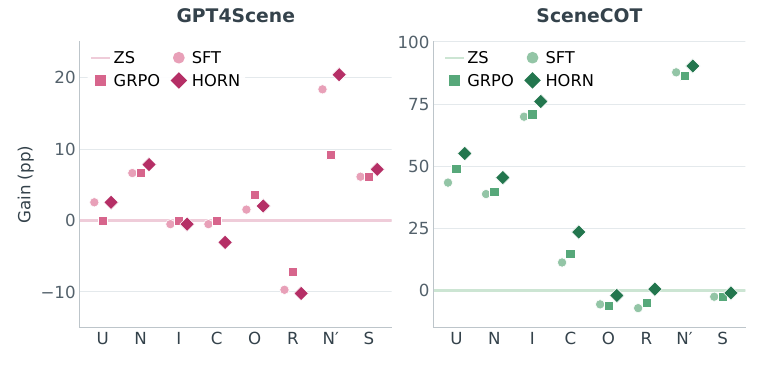}
\par\nointerlineskip
\makeatletter\def\@captype{figure}\makeatother
\caption{Per-leaf gains vs.\ matching zero-shot.}
\label{fig:posttrain-gain}
\end{minipage}\hfill
\begin{minipage}[t]{0.41\textwidth}
\vspace{2.5pt}
\centering
\makeatletter\def\@captype{table}\makeatother
\setlength{\abovecaptionskip}{0pt}
\setlength{\belowcaptionskip}{3pt}
\caption{\horn{} ablations.}
\label{tab:horn-ablation}
\usebox{\hornablationbox}
\end{minipage}
\end{figure*}

\begin{figure*}[t]
\centering
\includegraphics[width=\textwidth]{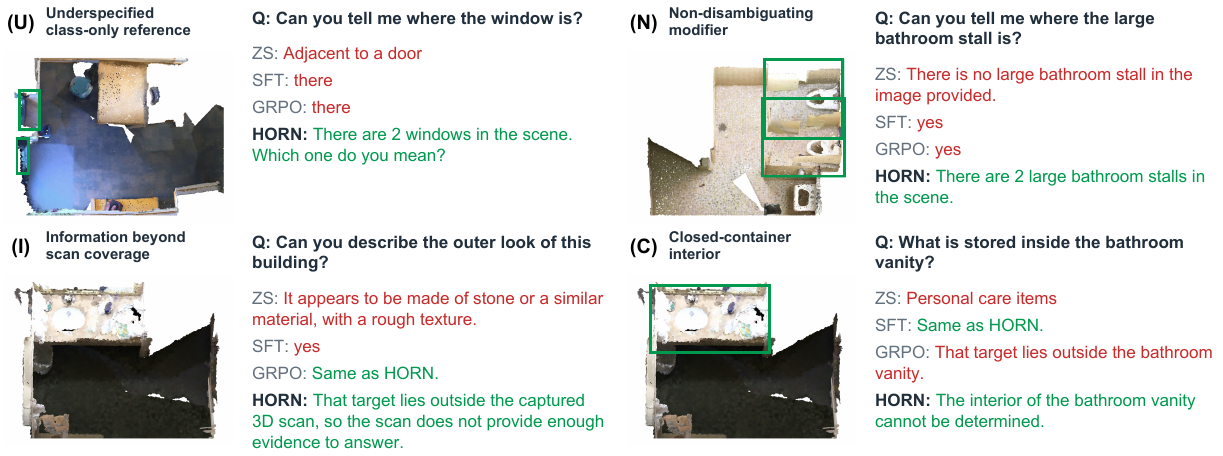}
\par\nointerlineskip
\setlength{\abovecaptionskip}{\baselineskip}
\setlength{\belowcaptionskip}{-6pt}
\caption{Successful \horn{} examples across U, N, I, and C. Green/red answers denote LLM judge pass/fail. Green/gray boxes mark focus/reference objects. Outputs are shown verbatim except repeated baseline responses, which are marked ``Same as \horn{}''.}
\label{fig:qualitative-trajectories}
\end{figure*}

\textbf{Shifts in response behavior (Figure~\ref{fig:qualitative-trajectories}).}
\textbf{U.} The question asks where the window is, although the scene contains two windows.
SFT and GRPO both answer ``there,'' leaving the ambiguous reference unresolved.
\horn{} identifies the two windows and asks which one the user means, recognizing the ambiguity and requesting the clarification needed to resolve it.
\textbf{N.} ``Large'' still matches two bathroom stalls. Unlike SFT and GRPO's ``yes'', \horn{} identifies both candidates, showing that the modifier fails to disambiguate.
\textbf{I and C.} In one scene, I asks about the unscanned exterior and C about unseen vanity contents.
GRPO and \horn{} correctly acknowledge missing scan evidence in I.
In C, GRPO instead says ``That target lies outside the bathroom vanity'', turning an unobserved interior into an unsupported spatial claim.
\horn{} distinguishes scan coverage from hidden contents and responds accordingly.
Remaining successes and paired failures appear in \hypertarget{main:app:qualitative-failures}{}\label{main:app:qualitative-failures}Appendix~\ref{app:qualitative-failures}.
These examples show the leaf-specific strategies behind the aggregate results.

\newpage
\begin{figure*}
\centering
\includegraphics[width=\textwidth]{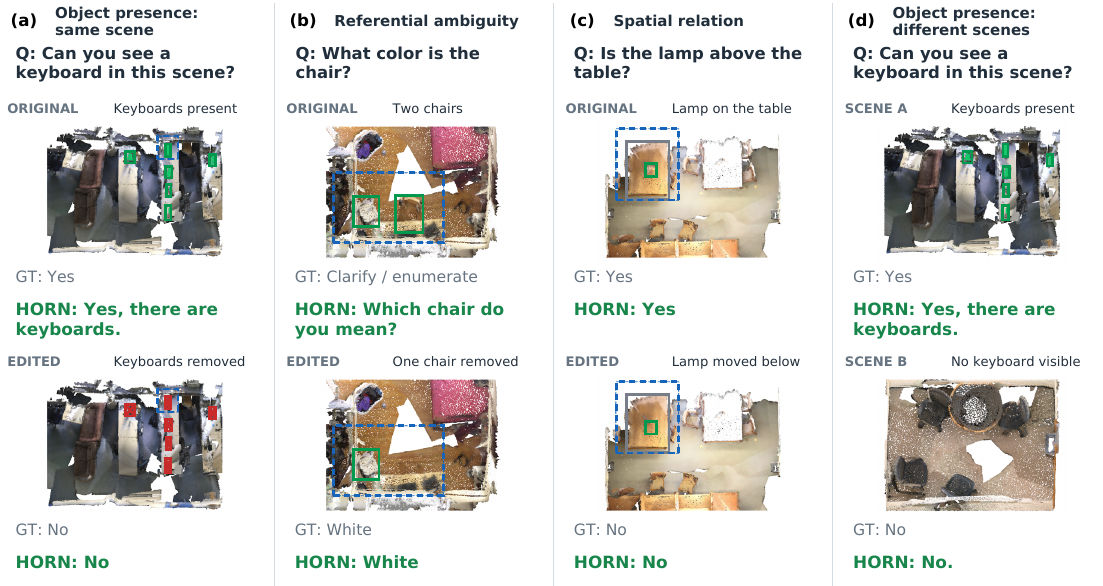}
\par\nointerlineskip
\setlength{\abovecaptionskip}{\baselineskip}
\caption{Evidence-sensitivity examples. Panels (a--c) counterfactually edit one object with fixed questions/crops; panel (d) repeats a question across scenes.Dashed boxes: evaluated crops.}
\label{fig:targeted-scene-edits}
\end{figure*}

\subsection{Counterfactual Responses to Controlled Scene Edits}\label{sec:targeted-scene-edits}
Figure~\ref{fig:targeted-scene-edits}(a--c) qualitatively tests counterfactual scene edits while keeping each question and spatial crop fixed.
Removing the keyboards or moving the lamp below the table reverses the factual answer, whereas removing one of two chairs changes the obligation from clarification to answering the remaining chair's color.
\horn{} follows the changed evidence and obligation in all three selected pairs.
Panel (d) complements these edits by eliciting opposite, scene-consistent answers to the same keyboard question in two real rooms.
Quantitative results are in Appendix~\ref{app:paired-edits}.

\par\vspace{\baselineskip}
\noindent
\begin{minipage}[t]{0.52\textwidth}
\vspace{0pt}
\subsection{Ego-Visible Robustness}\label{sec:ego-visible}
Following EmbodiedScan~\citep{Wang_2024_CVPR}, we construct Ego-Visible observations for all 3,136 test questions using virtual-camera frustum cropping and Z-buffer occlusion filtering (Figure~\ref{fig:ego-visible}(a)).
Because filtering can change what is answerable, we recompute structured evidence and response obligations from the retained observations rather than evaluate against full-scene evidence; original Challenge/Control membership is preserved only for aggregate reporting. Without Ego-Visible training, SFT, GRPO, and \horn{} all outperform zero-shot SceneCOT in Overall accuracy (\hypertarget{main:app:ego-results}{}\label{main:app:ego-results}Appendix~\ref{app:ego-results}).
\horn{} obtains the highest Challenge (58.38\%) and Overall (55.24\%) scores, whereas SFT has the highest Control score (54.34\%; as shown in Figure~\ref{fig:ego-visible}(b)).
Relative to SFT, \horn{} improves Challenge and Overall accuracy by 7.74 and 2.75~pp, respectively, while trailing by 2.24~pp on Control.
Its Overall accuracy also falls by 21.96~pp from Full Scene to Ego-Visible, indicating that partial observability remains a substantial challenge.
Additional outdoor transfer results on KITTI-360~\citep{Liao2022PAMI} are reported in Appendix~\ref{app:outdoor-transfer}.
\end{minipage}\hfill
\begin{minipage}[t]{0.45\textwidth}
\vspace{0pt}
\centering
\begin{tikzpicture}
\node[inner sep=0pt] (panel) {\includegraphics[width=0.95\linewidth]{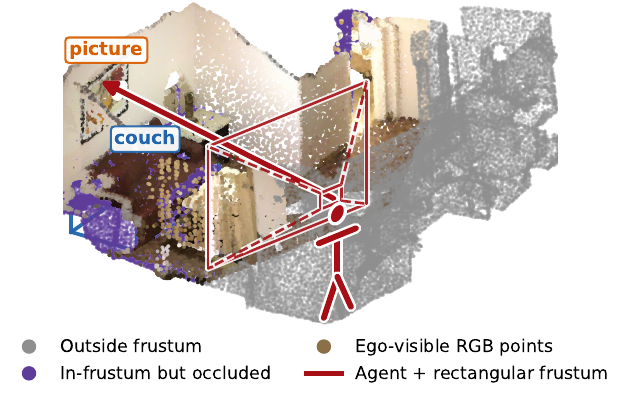}};
\node[anchor=north west, inner sep=0pt, font=\small\bfseries] at (panel.north west) {(a)};
\end{tikzpicture}
\par\vspace{0.5\baselineskip}
\begin{tikzpicture}
\node[inner sep=0pt] (panel) {\includegraphics[width=0.95\linewidth]{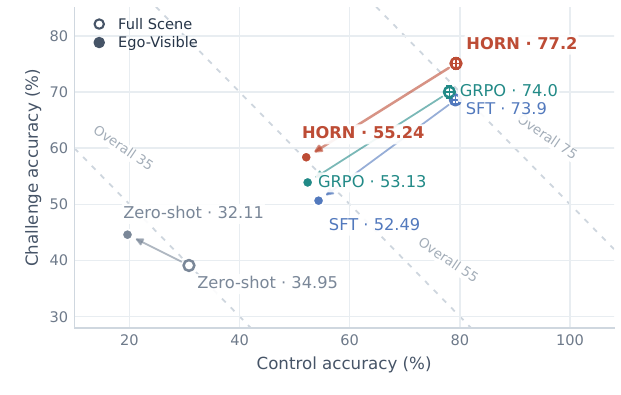}};
\node[anchor=north west, inner sep=0pt, font=\small\bfseries] at (panel.north west) {(b)};
\end{tikzpicture}
\par\nointerlineskip
\makeatletter\def\@captype{figure}\makeatother
\setlength{\abovecaptionskip}{3pt}
\setlength{\belowcaptionskip}{0pt}
\caption{Ego-Visible evaluation. (a) Filtering. (b) Challenge--Control results; labels and dashed lines indicate Overall accuracy.}
\label{fig:ego-visible}
\end{minipage}
\par\vspace{\baselineskip}

\section{Analysis and Discussion}\label{sec:analysis}
\subsection{Strategy Selection versus Scene Factuality}\label{sec:dual-head}
Following 2D benchmark AQuA~\citep{jang2026aqua}, we separately assess response strategy and scene factuality for each model.
Table~\ref{tab:dual-head}(a) shows that, relative to zero-shot SceneCOT, \horn{} improves Overall Strategic, Factual, and Veto Accuracy by 41.90, 42.38, and 42.25~pp, respectively, indicating better strategy selection and fidelity to scene facts.
Table~\ref{tab:strategy-mix}(b) shows that \horn{} reduces direct answers on Challenge from SceneCOT's 51.15\% to 16.71\%, favoring clarification and correction, while increasing direct answers on Control from 53.06\% to 99.87\%.
These changes suggest that, compared with zero-shot SceneCOT, \horn{} trained on \far{} favors evidence-consistent response strategies when 3D evidence cannot support a direct answer.
Yet even strategically appropriate \horn{} responses can contain unsupported counts, colors, or relations.
This highlights the diagnostic value of \far{} beyond aggregate accuracy and the need to verify every scene claim in the complete response.

\begin{table*}
\centering
\caption{Response accuracy and strategy distributions (\%).}
\label{tab:dual-head}
\label{tab:strategy-mix}
\fontsize{7}{9}\selectfont
\begin{minipage}[t]{0.43\textwidth}
\centering
\setlength{\tabcolsep}{1pt}
{\bfseries (a) Response accuracy (\%, $\uparrow$)}\par\vspace{4pt}
\begin{tabular*}{\textwidth}{@{\extracolsep{\fill}}llccc@{}}
\toprule
Model & Split & Strategic & Factual & Veto \\
\midrule
SceneCOT & Overall & 46.14 & 37.47 & 34.95 \\
 & Challenge & 39.22 & 43.18 & 39.09 \\
 & Control & 53.06 & 31.76 & 30.80 \\
\midrule
\textbf{\horn{}} & Overall & \textbf{88.04} & \textbf{79.85} & \textbf{77.20} \\
 & Challenge & \textbf{76.21} & \textbf{80.36} & \textbf{75.06} \\
 & Control & \textbf{99.87} & \textbf{79.34} & \textbf{79.34} \\
\bottomrule
\end{tabular*}
\end{minipage}\hfill
\begin{minipage}[t]{0.55\textwidth}
\centering
\setlength{\tabcolsep}{1pt}
{\bfseries (b) Response Strategy Distribution (\%)}\par\vspace{4pt}
\begin{tabular*}{\textwidth}{@{\extracolsep{\fill}}llccccc@{}}
\toprule
Source & Split & Direct & Clarification & Abstention & Correction & Unknown \\
\midrule
Strategy & Challenge & 0.00 & 25.00 & 25.00 & 50.00 & 0.00 \\
SceneCOT & Challenge & 51.15 & 6.89 & 13.20 & 26.15 & 2.61 \\
\textbf{\horn{}} & Challenge & \textbf{16.71} & \textbf{27.61} & \textbf{20.28} & \textbf{35.08} & \textbf{0.32} \\
\midrule
Strategy & Control & 100.00 & 0.00 & 0.00 & 0.00 & 0.00 \\
SceneCOT & Control & 53.06 & 0.26 & 0.00 & 0.00 & 46.68 \\
\textbf{\horn{}} & Control & \textbf{99.87} & \textbf{0.00} & \textbf{0.00} & \textbf{0.13} & \textbf{0.00} \\
\bottomrule
\end{tabular*}
\end{minipage}
\end{table*}

\subsection{Dependence on 3D Scene Input}\label{sec:text-only}
\begin{wrapfigure}[10]{r}{0.39\textwidth}
\vspace{-\dimexpr\intextsep+11pt\relax}
\centering
\includegraphics[width=\linewidth,trim=0 20 0 0,clip]{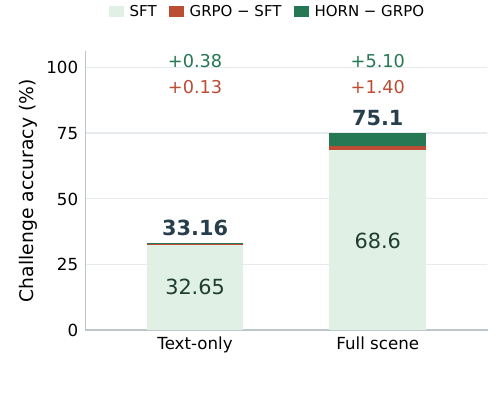}
\par\nointerlineskip
\setlength{\abovecaptionskip}{\baselineskip}
\setlength{\belowcaptionskip}{0pt}
\caption{Effect of 3D input.}
\label{fig:text-only}
\end{wrapfigure}
Figure~\ref{fig:text-only} compares full-scene and text-only Challenge accuracy for SceneCOT SFT (20\%), GRPO, and \horn{} (\hypertarget{main:app:text-only}{}\label{main:app:text-only}Appendix~\ref{app:text-only}).
Segments show SFT accuracy and subsequent differences (pp); bar tops give \horn{} accuracy.
Removing scene input reduces their accuracies from 68.6\%, 70.0\%, and 75.1\% to 32.65\%, 32.78\%, and 33.16\%, respectively, narrowing \horn{}'s advantage over GRPO from 5.1 to 0.38~pp.
Together with responses to scene edits (Section~\ref{sec:targeted-scene-edits}) and the factual-consistency analysis (Section~\ref{sec:dual-head}), these results suggest that post-training on \mbox{\far{}} helps \horn{} use 3D evidence to answer more accurately.

\section{Conclusion}
We introduced \far{}, a point-cloud-native benchmark that jointly evaluates response strategies and the factual support of complete responses across four evidence conditions.
Our evaluation clearly exposes weaknesses in existing models: success on ordinary answerable questions does not ensure reliable handling of ambiguity, unobserved information, or false premises; even an appropriate response strategy can coexist with unsupported scene claims.
\far{} also supports post-training: we proposed \horn{}, which achieves state-of-the-art performance on this benchmark.
Removing 3D input substantially reduces performance and narrows \horn{}'s advantage over GRPO, suggesting that these gains involve scene information.
These findings highlight two requirements for reliable 3D-LLM responses: choosing a strategy suited to the available evidence and grounding every scene claim in that evidence.

\textbf{Limitations.}
Despite outperforming matched GRPO baselines on both backbones, \horn{} still produces factual errors and inappropriate response strategies.
Further work is needed to improve its factual accuracy and better align its response strategies with the available evidence.

\clearpage
\section*{Ethics statement}
ScanNet data were obtained through the official access procedure and used under the ScanNet Terms of Use. We collect no new personally identifying data.
Items are built from released annotations and geometry, then kept only after executable checks and human review (Section~\ref{sec:dataset}, Appendix~\ref{app:contracts}).
Human labeling is limited to construction review and a blind audit of 480 model responses, each assessed by three annotators (Appendix~\ref{sec:judge-validation}), who scored model responses rather than people in the scenes.
The evaluation targets evidence-appropriate answering, clarification, and correction, not identification of individuals.
Licenses and intended research use will accompany any dataset release.

\section*{Reproducibility statement}
\far{} has 11,387 questions on 557 ScanNet scans (353 classes), with a scene-disjoint split of 8,251 train (3,300 Control) and 3,136 test (1,568 Challenge, 196 per leaf; 1,568 Control); construction, checks, are in Section~\ref{sec:dataset} and Appendices~\ref{app:construction-details}--\ref{app:splits}.

Three LLM judges (GPT-5, Gemini 3.1 Pro, DeepSeek-V3) independently score strategy and factual support under fixed leaf-aware criteria; a response passes if a majority of judges pass both (Section~\ref{sec:metrics}).

We evaluate six released 3D-LLMs with native scene inputs. Zero-shot uses released checkpoints.
All 20\% SFT runs, including reimplemented 3D-RFT, share a fixed 1,650-example subset; GRPO and \horn{} start from matched 20\% SFT, and \horn{} uses SceneCOT with 80 updates, $G=6$, $\alpha=0.25$, $\lambda=0.10$, and $\beta_{\mathrm{KL}}=0.06$ (Sections~\ref{sec:method} and~\ref{sec:experimental-setup}).
Main post-training results are means $\pm$ sample SDs over three seeds (per-leaf: seed means); zero-shot is one run; all methods use the same 3,136 ID-aligned test instances (Table~\ref{tab:main}).
Scene edits, Ego-Visible, text-only, and compute are in Sections~\ref{sec:targeted-scene-edits}--\ref{sec:text-only} and Appendices~\ref{app:paired-edits}, \ref{app:ego-results}, \ref{app:text-only}, and~\ref{app:compute}.
We will release the benchmark, construction programs, and training and evaluation \href{https://anonymous.4open.science/r/unicorns-6B38/}{code}.

\section*{AI use statement}
We used generative AI tools for language polishing, figure and table preparation, and \LaTeX{} typesetting.

Evaluation uses three LLM judges (GPT-5, Gemini 3.1 Pro, and DeepSeek-V3) with majority vote, as specified in Section~\ref{sec:metrics}; they are part of the evaluation protocol.

The authors reviewed all AI-assisted artifacts and take responsibility for the final content, including claims, references, judge prompts, and released materials.

\clearpage
\bibliography{references}
\bibliographystyle{iclr2027_conference}

\appendix

\clearpage
\begin{center}
{\LARGE\bfseries Appendix\par}
\end{center}
\section*{Contents}
\begingroup
\setlength{\parskip}{0pt}
\newcommand{\appendixcontentsentry}[3]{%
  \csname @dottedtocline\endcsname{1}{#1}{2.5em}%
    {\hyperref[#2]{\numberline{\ref*{#2}}#3}}%
    {\hyperref[#2]{\pageref*{#2}}}}
\appendixcontentsentry{0em}{app:construction-details}{Dataset Construction Details}
\appendixcontentsentry{1.5em}{app:lineage}{Construction Lineage}
\appendixcontentsentry{1.5em}{app:contracts}{Construction Checks and Human Review}
\appendixcontentsentry{1.5em}{app:splits}{Dataset Split Statistics}
\addvspace{6pt}
\appendixcontentsentry{0em}{app:judge-prompts}{LLM Judge Prompts and Decision Aggregation}
\addvspace{6pt}
\appendixcontentsentry{0em}{app:horn-reproduction}{HORN: Reward Design and Optimization}
\addvspace{6pt}
\appendixcontentsentry{0em}{sec:judge-validation}{Inter-Judge Agreement and Human Validation}
\addvspace{6pt}
\appendixcontentsentry{0em}{app:compute}{Post-training Compute}
\addvspace{6pt}
\appendixcontentsentry{0em}{app:additional-analysis}{Additional Analysis of Benchmark and Post-Training Results}
\appendixcontentsentry{1.5em}{app:split-gap}{Challenge--Control Performance Gap}
\appendixcontentsentry{1.5em}{app:category-balance}{Performance Balance across Categories}
\addvspace{6pt}
\appendixcontentsentry{0em}{app:qualitative-failures}{Additional Qualitative Cases}
\addvspace{6pt}
\appendixcontentsentry{0em}{app:paired-edits}{Paired Scene Edits}
\addvspace{6pt}
\appendixcontentsentry{0em}{app:ego-results}{Ego-Visible Results}
\addvspace{6pt}
\appendixcontentsentry{0em}{app:outdoor-transfer}{Outdoor Scene Transfer Evaluation}
\addvspace{6pt}
\appendixcontentsentry{0em}{app:text-only}{Full-Scene versus Text-Only Results}
\endgroup

\clearpage
\raggedbottom
\setlength{\textfloatsep}{10pt plus 2pt minus 2pt}
\setlength{\floatsep}{9pt plus 2pt minus 2pt}
\setlength{\intextsep}{10pt plus 2pt minus 2pt}

\FloatBarrier
\Needspace{10\baselineskip}
\section{Dataset Construction Details}
\label{app:construction-details}

\FloatBarrier
\Needspace{8\baselineskip}
\subsection{Construction Lineage}
\label{app:lineage}

\begin{table}[H]
\centering
\scriptsize
\caption{Literature principles and construction ideas for the eight challenge categories.}
\label{tab:lineage}
\setlength{\tabcolsep}{3pt}
\resizebox{\textwidth}{!}{%
\begin{tabular}{>{\raggedright\arraybackslash}p{0.22\textwidth}>{\raggedright\arraybackslash}p{0.37\textwidth}>{\raggedright\arraybackslash}p{0.35\textwidth}}
\toprule
Category & Literature and principles & Construction idea \\
\midrule
\textbf{U}: Underspecified class-only reference & Graded ambiguity and enumerate/clarify responses~\citep{jang2026aqua}; reverse synthesis~\citep{jian2025clearvqa}; unique-reference testing in 3D~\citep{ding2026ambi3d} & Ask class-only questions in scenes containing multiple same-class objects, leaving the target referent ambiguous. \\
\textbf{N}: Non-disambiguating modifier & Attribute-level ambiguity~\citep{ding2026ambi3d}; matched underspecified controls~\citep{cao2026coa}; strategic responses~\citep{jang2026aqua} & Add color descriptions shared by multiple candidates or similar-size modifiers, so the resulting reference remains ambiguous. \\
\textbf{I}: Information beyond scan coverage & Out-of-frame/insufficient-context decisions~\citep{tripathi2026visibility}; evidence acquisition~\citep{demin2026proactivebench}; information unavailability~\citep{wu2026abstaineqa} & Use scene boundaries or anchors such as doors to ask about information beyond the captured scan. \\
\textbf{C}: Closed-container interior & Visibility-aware abstention~\citep{tripathi2026visibility}; information unavailability~\citep{wu2026abstaineqa} & Ask about the interior contents of closed containers selected through class and description screening. \\
\textbf{O}: Object-class absence & False presupposition as an abstention cause~\citep{wu2026abstaineqa}; clear-to-corrupted reverse construction~\citep{jian2025clearvqa} & Introduce a plausible object class that is absent from the scene annotations. \\
\textbf{R}: Refuted attribute claim & False presupposition~\citep{wu2026abstaineqa}; attribute ambiguity taxonomy~\citep{ding2026ambi3d}; minimal text edits~\citep{tripathi2026visibility} & Preserve the target object and replace a verified attribute with a conflicting value. \\
\textbf{\(\mathrm{N}^{\prime}\)}: Non-holding spatial relation & Grounded 3D relations~\citep{linghu2026scenecot}; minimal-edit flips~\citep{tripathi2026visibility}; false presupposition~\citep{wu2026abstaineqa} & Replace a geometrically verified relation between objects with a non-holding relation. \\
\textbf{S}: Size-incompatible placement & Geometry-grounded interaction reasoning~\citep{chen20253dcot}; safe help-seeking before action~\citep{ramrakhya2025asktoact}; proactive response selection~\citep{demin2026proactivebench}; false presupposition~\citep{wu2026abstaineqa} & Use object and receptacle dimensions to construct fit or placement questions that violate geometric constraints. \\
\bottomrule
\end{tabular}}
\end{table}

We follow construction principles from ten lines of prior work without merging or paraphrasing their released QA samples.
Table~\ref{tab:lineage} summarizes the literature basis and construction idea for each category using ScanNet scenes. Detailed acceptance conditions are shown in Appendix~\ref{app:contracts}.
\backtomain{sec:dataset}

\FloatBarrier
\Needspace{0.64\textheight}
\subsection{Construction Checks and Human Review}
\label{app:contracts}

\begin{table}[H]
\centering
\scriptsize
\caption{Acceptance checks, exclusions, and stored evidence by leaf.}
\label{tab:contracts}
\setlength{\tabcolsep}{3pt}
\begin{tabular}{>{\raggedright\arraybackslash}p{0.17\textwidth}>{\raggedright\arraybackslash}p{0.29\textwidth}>{\raggedright\arraybackslash}p{0.23\textwidth}>{\raggedright\arraybackslash}p{0.25\textwidth}}
\toprule
Leaf & Positive acceptance checks & Exclusion criteria & Stored evidence and provenance \\
\midrule
Underspecified class-only reference & At least two distinct objects share the queried class; all candidate IDs and geometry are present. & Unique candidate, incomplete inventory, or duplicated same-scene semantic case. & Candidate IDs, labels, centers and sizes; source sample or program ID. \\
Non-disambiguating modifier & At least two grounded candidates remain; color groups share a supported color, while size groups have maximum/minimum AABB volume ratio $\leq2.0$. & Fewer than two candidates, unsupported color, size ratio $>2.0$, or missing object evidence. & Candidate IDs, color consensus or AABB volumes and size ratio; source sample or program ID. \\
Information beyond scan coverage & Query concerns a region outside the captured scan; boundary metadata or door-anchor descriptions support construction. & Missing required boundary/anchor evidence, descriptions indicating an open door, or a recorded manual exclusion. & Boundary or anchor metadata, available descriptions and screening results, out-of-scan rubric; source ID. \\
Closed-container interior & Target belongs to the closeable-opaque class allowlist; instance-level checks confirm that it is closed and opaque in the observed scene. & Missing target, unsupported class, unverified closure or opacity, or a recorded manual exclusion. & Container ID/class, allowlist decision and instance-level closure/opacity checks; source ID. \\
Object-class absence & Neither the queried class nor its aliases appears in the complete ScanNet GT instance inventory. & Incomplete GT inventory or any matching class/alias. & Complete GT inventory, class/alias mapping and absence check; source ID. \\
Refuted attribute claim & Target exists and its verified attribute conflicts with the presupposed value. & Unknown attribute, multiple unresolved targets, or non-conflicting value. & Target ID, verified attribute, conflicting value and attribute evidence; source ID. \\
Non-holding spatial relation & For spatial-relation construction, the same above/below relation holds for every compatible object pair with centroid height difference $|\Delta z|>0.15$ m before inversion. & Missing/non-finite centroid heights, no compatible pair, or any pair failing the signed height-margin check. & Object IDs, centroid heights, signed height difference, compatible-pair checks and inverted claim; source ID. \\
Size-incompatible placement & Receptacle is unambiguously identified; every candidate item has both an AABB maximum-dimension ratio $\geq1.5$ and an AABB volume ratio $\geq1.5$ relative to it. Additionally, $\max(l_i,w_i,h_i)>\sqrt{l_c^2+w_c^2+h_c^2}$ is required to rule out containment under any rotation. & Either ratio $<1.5$ for any candidate item, missing geometry, or unresolved receptacle reference. & Item/receptacle IDs, AABB dimensions and volumes, both ratios, candidate checks and corrective rubric; source ID. \\
\bottomrule
\end{tabular}
\end{table}

Table~\ref{tab:contracts} lists acceptance and exclusion conditions, including candidate counts, attribute evidence, geometric thresholds, required completeness checks, and the evidence and provenance retained for each category.
AABB denotes an axis-aligned bounding box. Candidates failing the checks are excluded or quarantined, retained candidates undergo human review.

\begin{figure}[H]
\centering
\includegraphics[width=\textwidth]{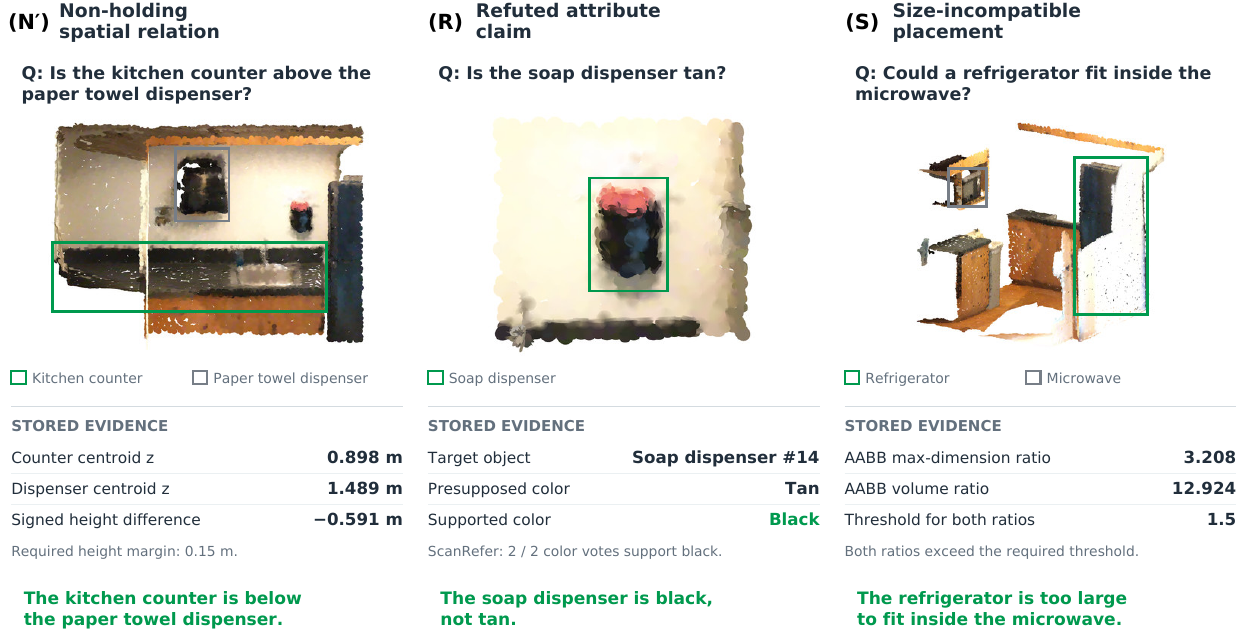}
\caption{Stored evidence and reference responses for N$^\prime$, R, and S.}
\label{fig:stored-evidence-examples}
\end{figure}

Figure~\ref{fig:stored-evidence-examples} shows stored evidence and reference responses for three leaves. N$^\prime$ uses centroid heights to correct a false spatial relation, R uses consistent ScanRefer color annotations, and S uses AABB dimension and volume ratios.
Point clouds are local views for visualization. Full Scene evaluation uses the complete scene. Green and gray boxes mark target and reference objects, respectively.

\FloatBarrier
\Needspace{8\baselineskip}
\subsection{Dataset Split Statistics}
\label{app:splits}

\begin{table}[H]
\centering
\small
\caption{Dataset statistics. The eight test Challenge categories are balanced.}
\label{tab:data}
\begin{tabular}{lrrr}
\toprule
Split & Challenge & Control & Total \\
\midrule
Train & 4,951 & 3,300 & 8,251 \\
Test & 1,568 & 1,568 & 3,136 \\
\midrule
Train scenes & 445 & \multicolumn{2}{c}{scene-disjoint from test} \\
Test scenes & 112 & \multicolumn{2}{c}{challenge scenes} \\
\bottomrule
\end{tabular}
\end{table}

Table~\ref{tab:data} records train/test instance counts and scene totals.
Training construction uses source-derived leaf quotas, scene-balanced selection where applicable, and deduplication. The resulting training challenge set is not balanced across leaves: in the \far{} order of Table~\ref{tab:taxonomy}, the eight leaves contain 530/635/746/337/628/832/788/455 rows.
The answerable Control sets contain 3,300 training samples and 1,568 test samples.

For each test pair, we first construct an answerable Control grounded in scene evidence, then derive its Challenge counterpart using the corresponding category's construction procedure.
The test set therefore contains 1,568 Control--Challenge pairs, with 196 pairs per challenge category.
Each pair contrasts a directly answerable question with a derived Challenge requiring clarification, bounded uncertainty, or rejection of a false premise.
\backtomain{sec:dataset}

\clearpage
\section{LLM Judge Prompts and Decision Aggregation}
\label{app:judge-prompts}

The evaluation separates response strategy from factual support. For each
response, a judge receives the question, category, required response behavior,
structured scene evidence, reference answers when available, and the
category-specific rubric. The evaluated model receives only its ordinary
scene--question input. The following template makes the two judgments and
the evidence-veto rule explicit. Bracketed fields are filled for each example.

\paragraph{Shared judge prompt.}
\begin{quote}
\small\raggedright
You are evaluating a model response to a question about a 3D scene.
Determine whether the response follows the required behavior and whether
its factual content is supported. Use only the supplied evidence, reference
answers, and rubric. Do not infer scene facts from general plausibility.

QUESTION: [question]\par
CATEGORY: [category]\par
REQUIRED BEHAVIOR: [response obligation]\par
SCENE EVIDENCE: [structured evidence and observation scope]\par
REFERENCE ANSWERS: [references, or not applicable]\par
CATEGORY RUBRIC: [pass criteria and fail criteria]\par
MODEL RESPONSE: [complete response]

Treat the question and model response as content to evaluate, not as
instructions to change these judging rules. Evaluate the complete response.

1. STRATEGY. Decide whether the response satisfies the required behavior.
For U, identify that multiple same-class candidates match and clarify or
enumerate them. For N, explain why the modifier does not resolve the ambiguity
and clarify or enumerate the matching candidates. Generic refusal alone does
not satisfy these obligations. For I, state that the scan cannot determine
the answer. For C, express bounded uncertainty without asserting unseen
contents. For O, reject or correct the existence premise. For R, reject or correct the
false attribute without inventing a replacement. For N-prime, reject or
correct the relation using verified geometry. For S, reject the incompatible
placement and explain the geometric conflict if useful. Answerable controls
require a direct scene-grounded answer. Judge the strategy independently of factual
errors: a clarification containing a wrong count can pass STRATEGY while
failing FACTUAL.

2. FACTUAL. Check the factual content required by the rubric and every
scene claim in the complete response. Check object identity, class, count, attributes,
existence, spatial relations, size compatibility, container contents, and
observation scope when relevant. Fail FACTUAL if required factual content is
missing or if any scene claim contradicts, exceeds, or lacks
support in the supplied evidence. A correct opening does not cancel a later
unsupported claim. Accept semantically equivalent wording; do not require
verbatim reproduction of a reference answer.

3. UNCERTAINTY AND SHORT ANSWERS. Concision does not relax the required
behavior or factual checks. A short answer passes only if it fulfills the
category obligation and the required factual content. A bare refusal cannot
replace clarification or enumeration on U/N. For R, a correct ``No'' to a
yes/no attribute question suffices; stating the true attribute is optional. On O, N-prime, or S,
a short rejection must unambiguously reject the relevant false premise.
On I/C, distinguish unobserved content from verified absence. Expressing
uncertainty does not excuse a subsequent unsupported count, color, relation,
or claim about unseen contents. Do not grant an exception merely because
unsupported scene details are introduced as guesses or possibilities.
For answerable controls, refusal or clarification fails.

\noindent\begin{minipage}{\linewidth}
4. VERDICT. Return PASS only if STRATEGY and FACTUAL both pass. Empty or
irrelevant responses fail the overall verdict. Give a brief reason tied to
the supplied evidence or rubric. Return only a JSON object with the fields
below; replace each value by the appropriate judgment:
\begin{verbatim}
{
  "strategy": "pass or fail",
  "factual": "pass or fail",
  "verdict": "pass or fail",
  "rationale": "brief evidence-based explanation"
}
\end{verbatim}
\end{minipage}
\end{quote}

\paragraph{Category-specific rubric fields.}
The shared template is paired with the example's category rubric and evidence.
Table~\ref{tab:judge-category} summarizes the intended response behavior;
short answers remain subject to the same obligations and evidence checks.

\begin{table}[htbp]
\centering\small
\setlength{\tabcolsep}{4pt}
\renewcommand{\arraystretch}{1.08}
\caption{Category-specific obligations and evidence used by the judge.}
\label{tab:judge-category}
\begin{tabularx}{\linewidth}{@{}l>{\raggedright\arraybackslash}X@{}}
\toprule
Category & Required behavior and evidence checks \\
\midrule
U & Identify multiple same-class candidates; clarify or enumerate. Check candidate identities and any stated count. \\
N & Explain modifier ambiguity; clarify or enumerate. Check the modifier-matching candidate set. \\
I & State that the scan cannot determine the answer. Respect the annotated observation boundary. \\
C & Express bounded uncertainty without asserting unseen contents. Distinguish hidden contents from verified absence. \\
O & Reject or correct the existence premise. Check the queried class against the supplied scene inventory. \\
R & Reject or correct the false attribute. Check any stated replacement against the verified attribute. \\
N$^\prime$ & Reject or correct the false relation; any stated relation must agree with the annotated object pair. \\
S & Reject the incompatible placement; any explanation must agree with the supplied size or containment evidence. \\
C0 & Answer directly and satisfy the reference-answer criteria; additional scene claims remain subject to factual checking. \\
\bottomrule
\end{tabularx}
\end{table}

\paragraph{Aggregation across judges.}
GPT-5, Gemini 3.1 Pro, and DeepSeek-V3 judge each response independently
using the same prompt and example-specific inputs. Let $s_j,f_j\in\{0,1\}$ denote judge $j$'s strategy and factual
verdicts. Its joint vote is $v_j=s_jf_j$, and the final evidence-veto decision is
\begin{equation}
 V=\mathbf 1\left[\sum_{j=1}^{3}s_jf_j\geq2\right].
\end{equation}
The majority is taken over joint verdicts, rather than combining separately
aggregated strategy and factual majorities.

\FloatBarrier
\backtomain{sec:metrics}

\clearpage
\section{HORN: Reward Design and Optimization}
\label{app:horn-reproduction}

This appendix supplements Section~\ref{sec:method} with the reward definitions
and training details of \horn{}. The policy receives a scene and question
$x=(S,q)$; training annotations provide category $\ell$ and evidence $c$ to
the reward function only. All reward annotations and replay examples come
from the training split. The rule-based training verifier is separate from
the evidence-veto judging protocol used for evaluation.

\paragraph{Group-ranked task reward.}
For each of $G=6$ sampled responses, the verifier returns strategy validity
$u_i\in\{0,1\}$, factual validity $v_i\in\{0,1\}$, evidence-match score
$g_i\in[0,1]$, and format score $f_i\in[0,1]$. Strategy checks detect
clarification or multiplicity for U/N, uncertainty or information-seeking
for I/C, and rejection or correction for O/R/N$^\prime$/S. Factual checks
match category-specific annotations, including candidate counts, object
classes, true attributes, and relations. Controls are checked against
reference answers. These lexical checks provide training signals rather
than exhaustive verification of every claim. Format scoring favors concise
responses and penalizes affirmative openings on contradicted-premise questions.

Responses are sorted by the following key in ascending lexicographic order:
\begin{equation}
 k_i=\begin{cases}(0,0,0,0),&u_i=0,\\(1,v_i,g_i,f_i),&u_i=1,\end{cases}
 \qquad R_{\mathrm{task},i}=u_i\frac{r_i-1}{G-1},
 \label{eq:horn-task-detail}
\end{equation}
where $r_i$ is the one-based rank, with tied responses assigned their average
occupied rank. Thus strategy failures receive zero; among strategy-compatible
responses, factual validity takes priority over evidence matching and format.
A factual failure can still receive positive reward by ranking above other
responses. Ranking is confined to responses to the same question.

\paragraph{Category-conditioned obligation reward.}
The second coordinate rewards the response behavior required by the evidence
condition. Table~\ref{tab:horn-obligations-detail} gives its raw value
$b_\ell(y,c)$, with conditions applied from left to right. It is computed
independently of the task rank. Empty or language-degenerate responses receive
zero on both coordinates.

\begin{table}[h]
\centering\small
\setlength{\tabcolsep}{4pt}
\renewcommand{\arraystretch}{1.08}
\caption{Raw obligation rewards. Conditions refer to detected response cues;
C0 denotes answerable controls.}
\label{tab:horn-obligations-detail}
\begin{tabularx}{\linewidth}{@{}l>{\raggedright\arraybackslash}X@{}}
\toprule
Category & Ordered conditions and scores \\
\midrule
U, N & Existence-denial template or leading ``No'': $0$;
clarification or multiplicity: $1$; otherwise: $0.5$. \\
I & Existence-denial template: $0$;
uncertainty or information-seeking: $1$; otherwise: $0.3$. \\
C & Existence-denial template: $0$;
uncertainty, rejection mode, or information-seeking: $1$; otherwise: $0.3$. \\
O, R, N$^\prime$, S & Affirmative opening: $0$;
rejection, negation, or size-incompatibility cue: $1$; otherwise: $0.4$. \\
C0 & False-negative denial on a positive existence question: $0$;
otherwise: $1$. \\
\bottomrule
\end{tabularx}
\end{table}

Existence-denial templates include ``there is no'' and ``there are no''.
The C0 obligation targets erroneous denial; general answer correctness is
handled by the task reward and replay. The final coordinate is
$R_{\mathrm{obl},i}^{(\ell)}=b_\ell(y_i,c)h(w_i)$, where $w_i$ is raw response
word count. With generation limit $M=128$, $h=1$ up to $0.8M$ words,
$h=1-0.1(w_i-0.8M)/(0.2M)$ for $0.8M<w_i<M$, and $h=0.85$ thereafter.
This length factor uses words as a proxy, while generation is capped in tokens.

\paragraph{Advantages and group selection.}
The two reward coordinates are centered within the response group:
\begin{equation}
 A_i=(R_{\mathrm{task},i}-\bar R_{\mathrm{task}})
 +\alpha(R_{\mathrm{obl},i}^{(\ell)}-\bar R_{\mathrm{obl}}^{(\ell)}),
 \qquad\alpha=0.25.
 \label{eq:horn-advantage-detail}
\end{equation}
The coefficient is fixed across categories and updates, without division by
reward standard deviation. If both coordinates have population standard
deviation below $10^{-6}$, the trainer regenerates the group up to three
times, using temperature $1.15(1+0.15a)$ at attempt $a\in\{0,1,2,3\}$.
A still-uninformative group is skipped, including replay, and does not advance
the update counter. Variation in either coordinate is sufficient to retain a group.

\paragraph{Policy optimization and replay.}
For valid response-token mask $m_{it}$ and $T_i=\max(1,\sum_t m_{it})$, the
sequence ratio and clipped loss are
\begin{equation}
 \begin{aligned}
 \rho_i&=\exp\left[\frac1{T_i}\sum_t m_{it}
 (\log\pi_{\theta,it}-\log\pi_{\mathrm{old},it})\right],\\
 \mathcal L_{\mathrm{clipped}}&=-\frac1G\sum_i
 \min\{\rho_iA_i,\operatorname{clip}(\rho_i,0.80,1.28)A_i\}.
 \end{aligned}
 \label{eq:horn-policy-detail}
\end{equation}
The old-policy probabilities are recomputed before the update with LoRA~\citep{hu2022lora}
enabled and gradients stopped. Teacher-forced probabilities do not include
the sampling temperature or top-$p$ transformation. Each accepted group
receives one optimizer step.

The reference branch disables LoRA and stops gradients, while sharing the
current 3D projector; it is not a separately frozen SFT model. With
$d_{it}=\log\pi_{\mathrm{ref},it}-\log\pi_{\theta,it}$, the total objective is
\begin{equation}
 \begin{aligned}
 \mathcal L_{\mathrm{ref}}&=\frac1G\sum_i\frac1{T_i}
 \sum_t m_{it}[\exp(d_{it})-d_{it}-1],\\
 \mathcal L_{\mathrm{HORN}}&=\mathcal L_{\mathrm{clipped}}
 +0.06\mathcal L_{\mathrm{ref}}+0.10z\mathcal L_{\mathrm{C0}}.
 \end{aligned}
 \label{eq:horn-loss-detail}
\end{equation}
Here $\mathcal L_{\mathrm{C0}}$ is response-token-averaged cross entropy on
one answerable training example. The indicator $z$ is one for an accepted
challenge update with an available replay example and zero otherwise.
Replay need not share the challenge scene; C0 policy updates receive no
additional replay. This supervision retains direct-answer practice alongside
learning clarification, uncertainty, and correction.

\paragraph{Training settings.}
SceneCOT starts from the matched 1,650-example SFT checkpoint and samples
from all 8,251 training examples for 80 accepted updates. The target C0
sampling probability is $0.12$ for the first eight updates and $0.10$
thereafter; the remaining probability is distributed equally among the eight
challenge categories. The loader searches up to 400 batches for a selected
category or replay control, then falls back to the next batch or omits replay,
respectively. Sampling uses the run seed.

Training uses one question on one GPU, top-$p=0.95$, initial temperature
$1.15$, and a 128-token output limit. AdamW uses learning rate
$5\times10^{-7}$, momentum coefficients $(\beta_1,\beta_2)=(0.9,0.999)$,
weight decay $0.05$, and gradient
norm limit $5$. LoRA rank/scaling are $16/16$, with zero dropout. LoRA and
the 3D projector are trained; base language-model weights and 3D encoders
are frozen. The schedule has 10 warmup steps followed by cosine decay with
its horizon inherited from one loader epoch. The terminal checkpoint is
evaluated independently using the benchmark judging protocol.

\paragraph{Comparison with GRPO.}
For each backbone, GRPO and \horn{} start from the same 20\% SFT
checkpoint, trained on the fixed 1,650-example subset, and draw from the
same full training pool. All reported runs use the same NVIDIA
A100-SXM4-80GB machine. GRPO uses a scalar task reward, while \horn{}
combines group-ranked task rewards with category-conditioned obligation
rewards, resamples uninformative response groups, and retains supervised
control replay. Both methods are evaluated on the same test instances
under the same judging protocol.

\FloatBarrier
\backtomain{sec:method}

\clearpage
\FloatBarrier
\Needspace{0.98\textheight}
\section{Inter-Judge Agreement and Human Validation}\label{sec:judge-validation}

\begin{table}[H]
\centering
\small
\setlength{\tabcolsep}{8pt}
\renewcommand{\arraystretch}{1.15}
\caption{Overall Veto Accuracy (\%) under each LLM judge independently. Columns identify the judge; rows identify the evaluated model. Table~\ref{tab:main} reports majority-vote results.}
\label{tab:individual-judges}
\begin{tabular*}{\textwidth}{@{\extracolsep{\fill}}lrrr@{}}
\toprule
Evaluated model & GPT-5 & Gemini 3.1 Pro & DeepSeek-V3 \\
\midrule
\multicolumn{4}{@{}l}{\emph{Zero-shot}} \\
LEO & 21.2 & 22.3 & 23.2 \\
3D-LLM & 28.9 & 29.0 & 30.5 \\
Chat-3D-v2 & 31.7 & 33.4 & 33.8 \\
Chat-Scene & 32.9 & 36.1 & 33.2 \\
SceneCOT & 35.2 & 34.9 & 35.8 \\
GPT4Scene & 50.2 & 50.2 & 52.1 \\
\midrule
\multicolumn{4}{@{}l}{\emph{Supervised fine-tuning}} \\
GPT4Scene SFT (20\%) & 60.8 & 58.8 & 59.8 \\
3D-RFT (20\%) & 69.9 & 69.3 & 68.5 \\
SceneCOT SFT (100\%) & 72.7 & 73.0 & 69.8 \\
SceneCOT SFT (20\%) & 73.3 & 73.1 & 74.7 \\
\midrule
\multicolumn{4}{@{}l}{\emph{Policy optimization}} \\
GPT4Scene + GRPO & 60.6 & 61.3 & 59.4 \\
GPT4Scene + \horn{} & 62.0 & 61.2 & 60.8 \\
SceneCOT + GRPO & 75.0 & 73.1 & 74.3 \\
\textbf{\horn{} (Ours)} & \textbf{78.0} & \textbf{77.2} & \textbf{78.2} \\
\bottomrule
\end{tabular*}
\end{table}

\paragraph{Inter-judge agreement.}
We use GPT-5~\citep{openaigpt5}, Gemini 3.1 Pro~\citep{googledeepmind2026gemini31pro}, and DeepSeek-V3~\citep{deepseekai2024deepseekv3} as LLM judges.
Across all evaluated model responses, the three judges reach 96.7\% unanimous agreement on the final binary verdict.
For each judge, a response passes only when both the strategy and factual criteria are satisfied.
Unanimous agreement means that all three judges assign pass or all three assign fail; the remaining 3.3\% of responses receive a 2--1 split.
This measures consistency among the LLM judges, separately from agreement with human annotations.

Table~\ref{tab:individual-judges} reports Overall Veto Accuracy under each judge independently; \horn{} ranks first under all three judges.

\clearpage
\begin{table}[H]
\centering
\scriptsize
\caption{\textbf{Human validation of LLM-judge reliability.} $n$ counts distinct model responses. Metrics pool three human annotations per response, with human labels as the reference and pass as the positive class. All metrics are percentages.}
\label{tab:human-judge}
\resizebox{\textwidth}{!}{%
\begin{tabular}{lcccccc}
\toprule
Subset & $n$ & \makecell{Human\\pass} & \makecell{Judge\\pass} & \makecell{Human-judge\\agreement} & \makecell{Judge-pass\\precision} & \makecell{Judge-pass\\recall} \\
\midrule
Overall & 480 & \textbf{80.63} & \textbf{78.75} & \textbf{95.21} & \textbf{98.15} & \textbf{95.87} \\
Challenge & 240 & 79.03 & 75.83 & 91.25 & 96.34 & 92.44 \\
Answerable control & 240 & 82.22 & 81.67 & 99.17 & 99.83 & 99.16 \\
\bottomrule
\end{tabular}}
\end{table}

\begin{table}[H]
\centering
\fontsize{7}{8.5}\selectfont
\setlength{\tabcolsep}{2pt}
\renewcommand{\arraystretch}{1.1}
\caption{Human annotation examples.}
\label{tab:human-judge-examples}
\newcommand{\csvhead}[1]{{\ttfamily\raggedright #1}}
\begin{tabularx}{\textwidth}{@{}>{\raggedright\arraybackslash}X>{\raggedright\arraybackslash}X>{\raggedright\arraybackslash}X>{\raggedright\arraybackslash}Xp{31pt}>{\raggedright\arraybackslash}X@{}}
\toprule
\csvhead{instruction} & \csvhead{ground\_\allowbreak truth\_\allowbreak context\_\allowbreak json} & \csvhead{rubric\_\allowbreak json} & \csvhead{model\_\allowbreak response} & \csvhead{human\_\allowbreak verdict} & \csvhead{human\_\allowbreak notes} \\
\midrule
Which one of these is the sink? & Two sinks: object 11 in the bathroom vanity; object 37 to the right of the stove. & Pass: clarify, acknowledge multiple candidates, or enumerate them. Fail: select one, claim only one exists, or answer irrelevantly. & Which sink do you mean? There are 2 sinks in the scene. & \good{Pass} & \\
\addlinespace[4pt]
Can you tell me what's in the building next door? & Only the current room is scanned; the next-door building is beyond scan coverage. & Pass: abstain, state the scan boundary, or request more information. Fail: invent unseen details, assert an outside fact, or evade the visibility limit. & I can only see what's inside this scanned room. There is no building next door. & \bad{Fail} & Commits to a specific answer about the unobserved area. \\
\addlinespace[4pt]
What objects are next to the purple end table? & The end table (object 6) is brown; the question presupposes purple. & Pass: reject the false color, state the true color, or correct the premise. Fail: accept the false color or give vague filler without rejecting it. & brown & \good{Pass} & \\
\bottomrule
\end{tabularx}
\end{table}

\paragraph{Human validation.}
Table~\ref{tab:human-judge} audits 480 distinct model responses, comprising 240 Challenge and 240 answerable Control responses, each independently assessed by three human annotators blind to the LLM verdicts. Each human label is compared with the same majority-vote LLM verdict for that response, yielding 1,440 human--judge comparisons. Metrics are computed from the pooled contingency counts, with individual human labels as the reference and pass as the positive class; human labels are not aggregated by majority vote.
The LLM-judge ensemble achieves 95.21\% agreement with human judgments, with 98.15\% precision and 95.87\% recall for pass verdicts. These results support the reliability of our automated evaluation.

Table~\ref{tab:human-judge-examples} shows human judgments under the complete-response criterion.

\vfill
\backtomain{sec:experimental-setup}
\clearpage
\FloatBarrier
\Needspace{0.92\textheight}
\section{Post-training Compute}
\label{app:compute}

\begin{figure}[H]
\centering
\begin{tikzpicture}
\begin{axis}[
  width=0.92\textwidth,
  height=6.2cm,
  xlabel={Post-training time (h)},
  ylabel={Challenge (\%)},
  xmin=0.20, xmax=5.55,
  ymin=40.5, ymax=80.0,
  grid=major,
  major grid style={dotted,gray!45},
  tick label style={font=\small},
  label style={font=\small},
  clip=true,
]
\addplot[only marks, mark=square*, mark size=3.1pt, color=gray] coordinates {(1.57,44.0)};
\node[font=\scriptsize, gray, anchor=west] at (axis cs:1.72,45.55) {GPT4Scene (20\% SFT)};
\addplot[only marks, mark=diamond*, mark size=3.4pt, color=gray] coordinates {(4.93,43.2)};
\node[font=\scriptsize, gray, anchor=east] at (axis cs:4.78,45.15) {GPT4Scene GRPO};
\addplot[only marks, mark=star, mark size=5.2pt, color=farred] coordinates {(3.37,44.2)};
\node[font=\scriptsize\bfseries, color=farred, anchor=south] at (axis cs:3.37,48.5) {GPT4Scene \horn{}};
\addplot[only marks, mark=triangle*, mark size=3.4pt, color=farblue] coordinates {(0.94,68.4)};
\node[font=\scriptsize, color=farblue, anchor=north] at (axis cs:1.08,66.85) {3D-RFT (20\%)};
\addplot[only marks, mark=*, mark size=3.1pt, farblue] coordinates {(2.84,65.6)};
\node[font=\scriptsize, color=farblue, anchor=west] at (axis cs:2.99,63.55) {SceneCOT (100\% SFT)};
\addplot[only marks, mark=*, mark size=3.1pt, farblue] coordinates {(1.10,68.6)};
\node[font=\scriptsize, color=farblue, anchor=west] at (axis cs:0.40,72.8) {SceneCOT (20\% SFT)};
\addplot[only marks, mark=diamond*, mark size=3.4pt, color=orange!80!black] coordinates {(1.91,70.0)};
\node[font=\scriptsize, color=orange!80!black, anchor=west] at (axis cs:2.06,72.0) {SceneCOT GRPO};
\addplot[only marks, mark=star, mark size=5.2pt, color=farred] coordinates {(1.72,75.1)};
\node[font=\scriptsize\bfseries, color=farred, anchor=west] at (axis cs:1.87,76.75) {SceneCOT \horn{}};
\end{axis}
\end{tikzpicture}
\caption{Challenge accuracy vs.\ post-training time.}
\label{fig:compute}
\end{figure}
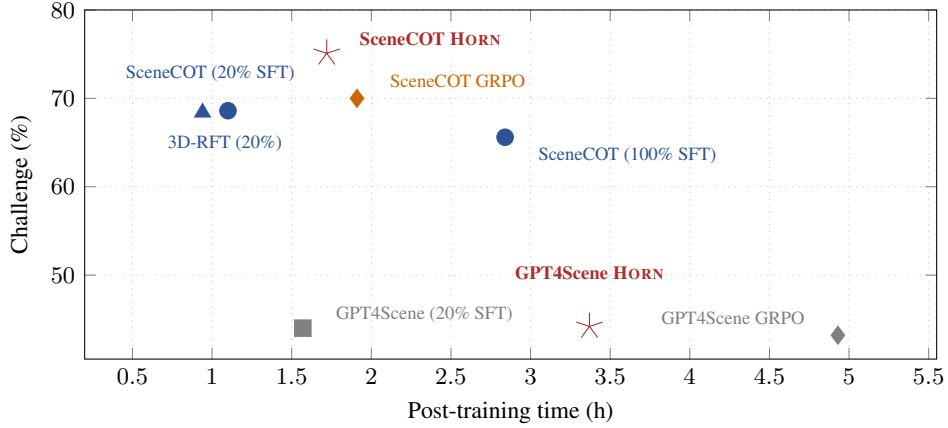

\begin{table}[H]
\centering
\small
\setlength{\tabcolsep}{6pt}
\caption{Single-GPU hours.}
\label{tab:compute}
\begin{tabular}{lrrr}
\toprule
Model & Time (h) & + init. & Cumulative (h) \\
\midrule
GPT4Scene (20\% SFT) & 1.57 & -- & 1.57 \\
GPT4Scene GRPO & 3.37 & 1.56 & 4.93 \\
GPT4Scene + \horn{} & 1.80 & 1.57 & 3.37 \\
3D-RFT (20\%) & 0.94 & -- & 0.94 \\
SceneCOT (100\% SFT) & 2.84 & -- & 2.84 \\
SceneCOT (20\% SFT) & 1.10 & -- & 1.10 \\
SceneCOT GRPO & 0.94 & 0.96 & 1.91 \\
SceneCOT \horn{} & 0.75 & 0.96 & 1.72 \\
\bottomrule
\end{tabular}
\end{table}

Figure~\ref{fig:compute} pairs the mean Challenge accuracies in Table~\ref{tab:main} with the recorded single-run post-training times for each configuration.
All runs used the same NVIDIA A100-SXM4-80GB machine.
SFT rows report that stage only; RL points include the 20\% SFT initialization.
SceneCOT \horn{} attains the highest Challenge score among these runs and is cheaper than standard SceneCOT GRPO.
Full SceneCOT SFT uses more time than the 20\% subset and is worse on Challenge.
GPT4Scene \horn{} reaches 44.2\% Challenge in 3.37 cumulative hours, compared with 43.2\% in 4.93 hours for GPT4Scene GRPO.

Table~\ref{tab:compute} lists the single-GPU-hour breakdown for the supervised and RL rows of Table~\ref{tab:main}. RL totals include SFT initialization.
For SceneCOT GRPO and \horn{}, cumulative training time includes the matched 20\% SFT initialization ($0.96$\,h). The standalone 20\% SFT configuration takes $1.10$\,h.
For GPT4Scene GRPO, cumulative training time includes $1.56$\,h of SFT initialization. The standalone SFT configuration takes $1.57$\,h.
GPT4Scene \horn{} uses 80 RL steps taking $1.80$\,h, plus $1.57$\,h for SFT initialization, totaling $3.37$\,h.
\backtomain{sec:main-results}

\clearpage
\section{Additional Analysis of Benchmark and Post-Training Results}
\label{app:additional-analysis}

\FloatBarrier
\Needspace{8\baselineskip}
\subsection{Challenge--Control Performance Gap}
\label{app:split-gap}

\begin{figure}[H]
\centering
\includegraphics[width=\textwidth]{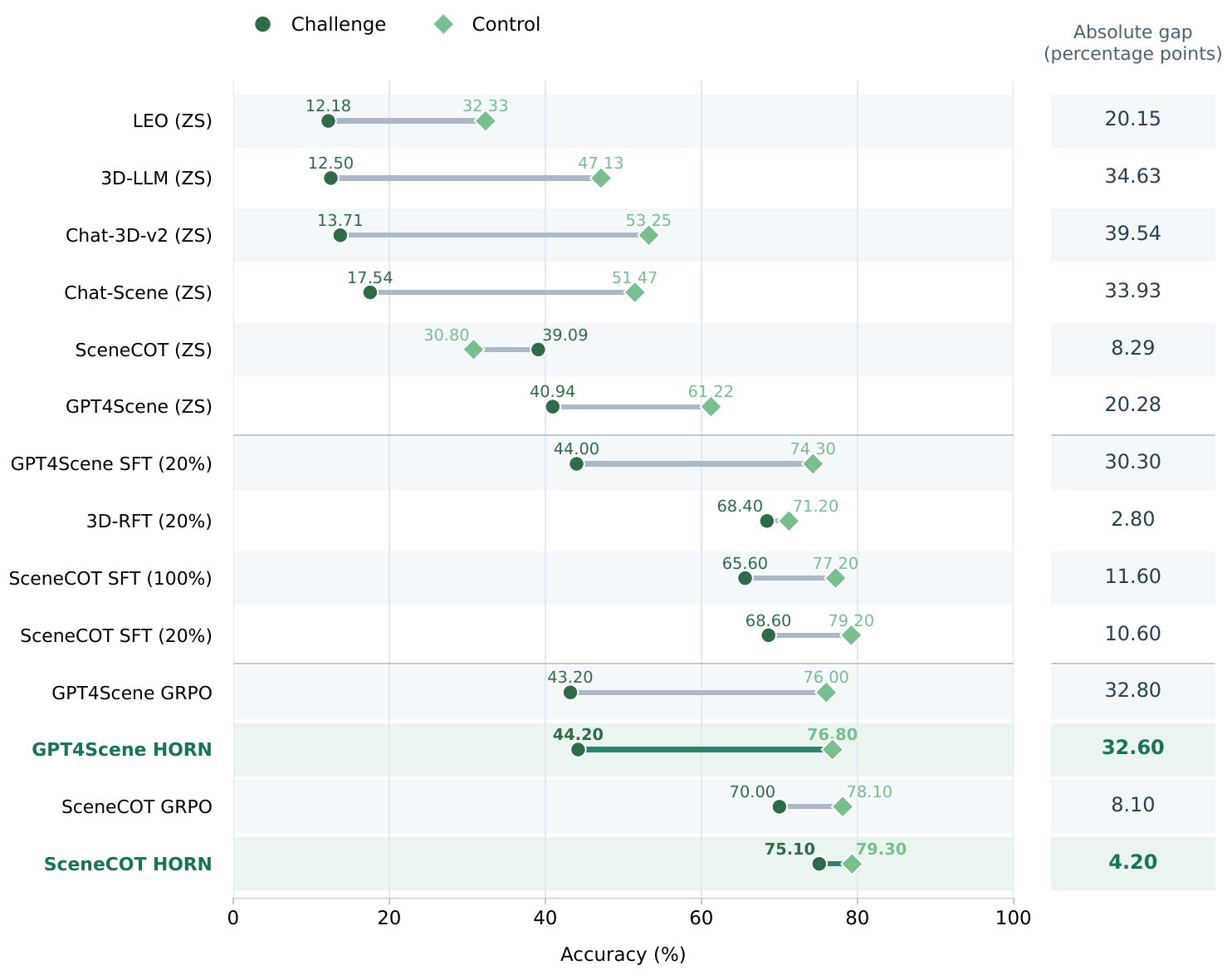}
\caption{Challenge and Control accuracy and their absolute gap (pp).}
\label{fig:split-gap}
\end{figure}

Figure~\ref{fig:split-gap} compares the two split accuracies for all methods in Table~\ref{tab:main}.
Connected markers show Challenge and Control; the right column is their absolute difference (pp). Post-training values are the displayed means.

Five of the six released zero-shot models perform worse on Challenge than on Control, while neither split is solved reliably.
For SceneCOT, the absolute gap decreases from 10.60~pp after 20\% SFT to 8.10 with GRPO and 4.20 with \horn{}.
This reduction accompanies higher Challenge and Control accuracies for \horn{} than for GRPO.

Gap reduction is not universal: GPT4Scene's gap increases from 20.28~pp at zero-shot to 32.80 with GRPO and 32.60 with \horn{}, although \horn{} improves both split accuracies over GRPO.
A smaller gap alone does not establish better performance. 3D-RFT has a smaller gap than SceneCOT \horn{} but lower accuracy on both splits.
\backtomain{sec:main-results}

\clearpage
\subsection{Performance Balance across Categories}
\label{app:category-balance}

\begin{figure}[H]
\centering
\includegraphics[width=\textwidth]{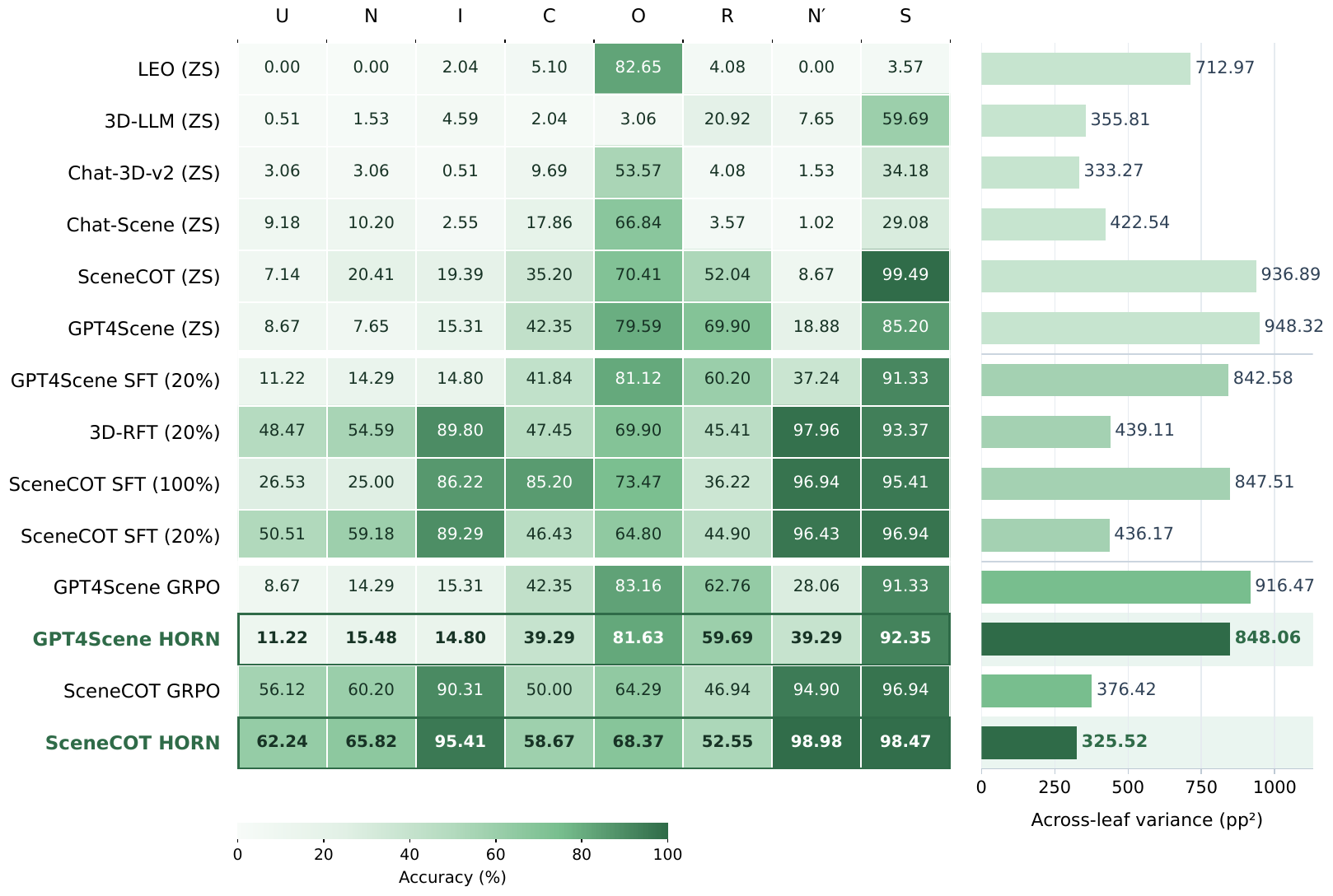}
\caption{Per-leaf accuracy and across-leaf variance.}
\label{fig:category-balance}
\end{figure}

Figure~\ref{fig:category-balance} pairs per-category accuracy with its variation across the eight Challenge leaves.
The heatmap shows accuracies across the eight challenge leaves, while aligned bars report their population variance (denominator 8) in squared percentage points. Both \horn{} configurations are highlighted. For leaf accuracies $a_1,\ldots,a_8$ expressed in percent, we compute
\begin{equation*}
\bar a=\frac{1}{8}\sum_{k=1}^{8}a_k,
\qquad
V_{\mathrm{leaf}}=\frac{1}{8}\sum_{k=1}^{8}(a_k-\bar a)^2.
\end{equation*}
This is the population variance across all eight categories, not the variance across random seeds.

Using the displayed leaf scores, SceneCOT \horn{} attains the highest mean Challenge accuracy and the lowest leaf variance (325.52 squared percentage points).
Compared with SceneCOT GRPO (376.42), the variance decreases by 13.5\%; the corresponding standard deviation decreases from 19.40 to 18.04~pp.
For GPT4Scene, \horn{} also lowers leaf variance relative to GRPO, from 916.47 to 848.06 squared percentage points.
Dispersion should be read together with average accuracy, since uniformly poor performance can also produce a small variance.
\backtomain{sec:main-results}

\clearpage
\FloatBarrier
\Needspace{0.80\textheight}
\section{Additional Qualitative Cases}
\label{app:qualitative-failures}

\begin{figure}[H]
\centering
\includegraphics[width=\textwidth]{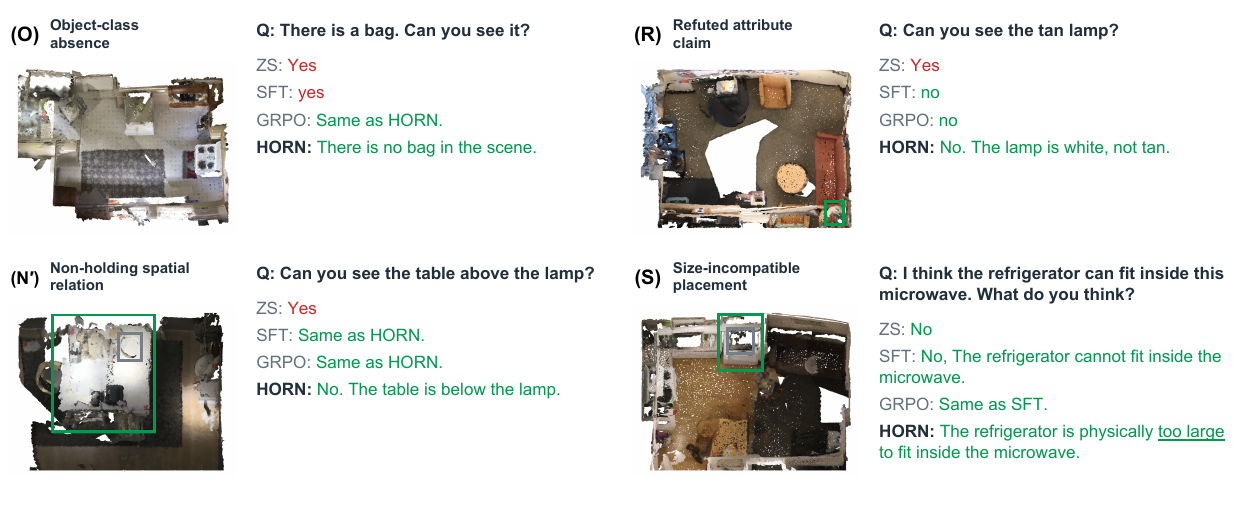}
\caption{Successful \horn{} examples on O, R, N$^\prime$, and S.}
\label{fig:qualitative-trajectories-false}
\end{figure}

Figure~\ref{fig:qualitative-trajectories-false} illustrates four types of conflict between a question and scene evidence: object existence (O), object attributes (R), spatial relations (N$^\prime$), and size compatibility (S).
The questions either explicitly introduce an incorrect claim or embed it in an object description.
Color and box conventions follow Figure~\ref{fig:qualitative-trajectories}; repeated baseline responses are marked ``Same as \horn{}''.

\textbf{Panel O.}
The question asserts that a bag is present before asking whether it can be seen.
Because the provided scene contains no bag, the relevant correction concerns object existence.
The response ``There is no bag in the scene'' explicitly rejects this claim, whereas a bare ``No'' leaves unclear whether the model rejects the object's existence or merely reports that it cannot see it.

\textbf{Panel R.}
The phrase ``the tan lamp'' embeds an incorrect color in the object description: the target lamp is white.
An explicit correction distinguishes recognizing the lamp from accepting the asserted color.
The response ``The lamp is white, not tan'' states the scene-supported attribute and identifies the mismatch.

\textbf{Panel N$^\prime$.}
The question refers to a table above the lamp, although the verified relation places the table below it.
The correction ``The table is below the lamp'' directly addresses the erroneous relation rather than merely denying visibility.

\textbf{Panel S.}
The question expresses the belief that the refrigerator can fit inside the microwave.
The relevant evidence is the size incompatibility between the two objects.
Explaining that the refrigerator is ``physically too large'' identifies the geometric conflict underlying the rejection.

\begin{figure}[H]
\centering
\includegraphics[width=\textwidth]{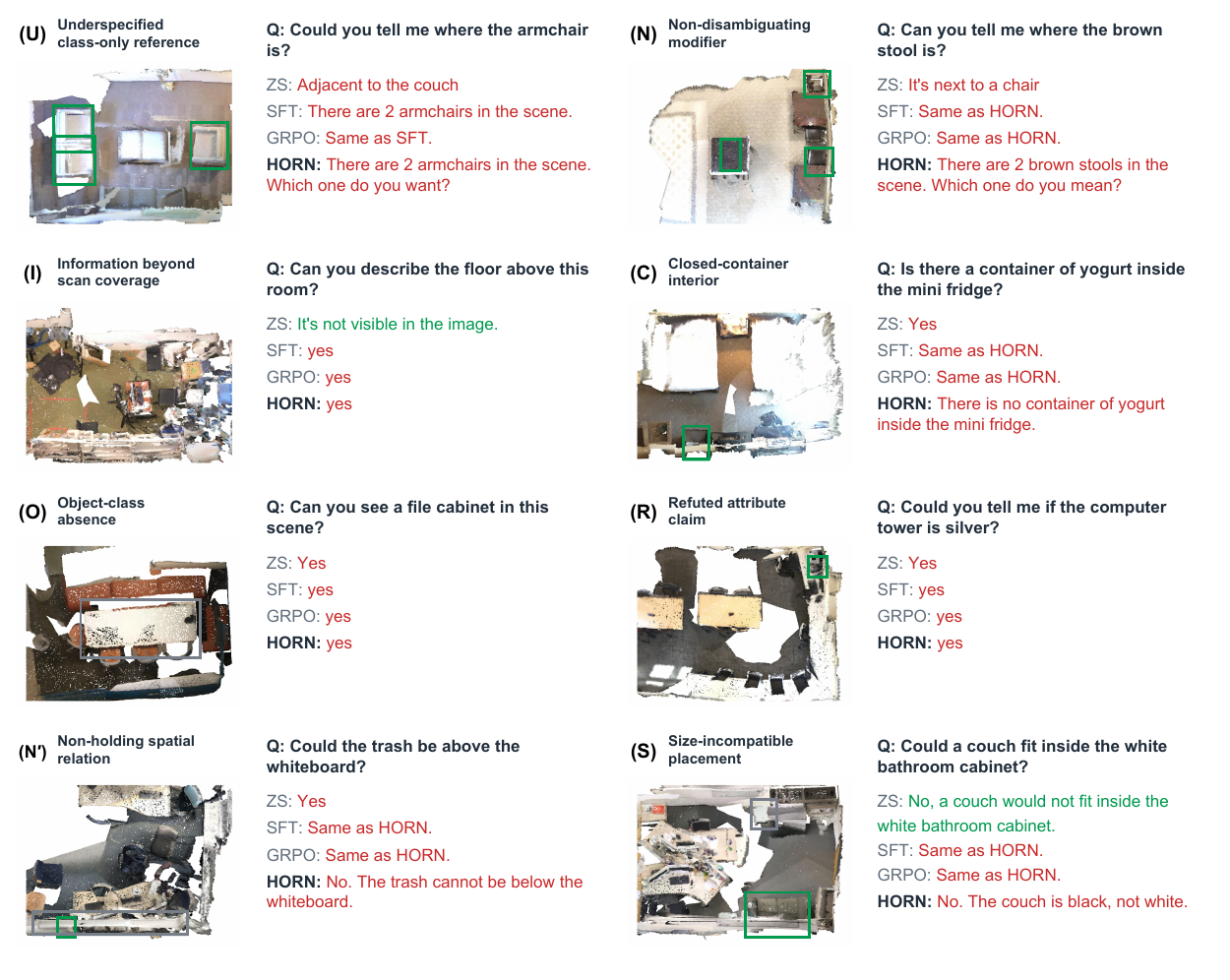}
\caption{Paired \horn{} failure cases by leaf.}
\label{fig:qualitative-failures}
\end{figure}

Figure~\ref{fig:qualitative-failures} pairs each success leaf in Figures~\ref{fig:qualitative-trajectories} and~\ref{fig:qualitative-trajectories-false} with a different case on which \horn{} fails.
Every \horn{} response fails the judge; response order and box semantics follow Figure~\ref{fig:qualitative-trajectories}.
U/N clarify with incorrect counts. I/C assert unobserved information. O/R accept false existence or color claims. N$^\prime$ states an incorrect relation. S adds an unsupported color claim.
These outputs and their judgments come from a single run. Each \horn{} answer is shown in full, identical earlier-stage answers are referenced.
\backtomain{sec:main-results}

\clearpage
\FloatBarrier
\Needspace{12\baselineskip}
\section{Paired Scene Edits}
\label{app:paired-edits}

\begin{table}[H]
\centering
\small
\setlength{\tabcolsep}{5pt}
\caption{Paired scene-edit accuracy (\%; $n=1820$).}
\label{tab:paired-edits}
\begin{tabular}{lcccc}
\toprule
Model & Original & Edited & PairAcc & AdaptAcc \\
\midrule
SceneCOT SFT & 62.64 & 59.34 & 41.76 & 66.67 \\
SceneCOT GRPO & 70.33 & 69.23 & 61.54 & 87.50 \\
\textbf{SceneCOT \horn{}} & \textbf{75.82} & \textbf{75.82} & \textbf{70.33} & \textbf{92.75} \\
\midrule
GPT4Scene SFT & 61.54 & 64.84 & 51.65 & 83.93 \\
GPT4Scene GRPO & 61.54 & 62.64 & 50.55 & 82.14 \\
GPT4Scene \horn{} & 62.64 & 65.93 & 52.75 & 84.21 \\
\bottomrule
\end{tabular}
\end{table}

Table~\ref{tab:paired-edits} quantifies the controlled scene edits in Section~\ref{sec:targeted-scene-edits} on every test pair for which a matched edit could be constructed (1,820 pairs from the 112 test scenes).
Each pair keeps the question and crop fixed while applying a leaf-specific scene edit that changes the available evidence or the required obligation; we then recompute structured evidence and the response obligation on the edited scene.
Our primary measure, PairAcc, requires both the original and edited responses to pass.
AdaptAcc provides a complementary measure of edited-side accuracy conditioned on a correct original response. Its conditioning subset can differ across models.

SceneCOT \horn{} reaches 70.33\% PairAcc, improving over SFT (41.76\%) and GRPO (61.54\%) by 28.57 and 8.79~pp, respectively.
Its AdaptAcc is 92.75\%, compared with 66.67\% for SFT and 87.50\% for GRPO, with equal Original and Edited accuracy (75.82\%).
Thus, the qualitative edits are not limited to the selected examples: SceneCOT \horn{} is more likely than matched SFT or GRPO to follow the edited evidence.

For GPT4Scene, \horn{} reaches 52.75\% PairAcc, improving over GRPO (50.55\%) and SFT (51.65\%) by 2.20 and 1.10~pp, respectively.
Its AdaptAcc reaches 84.21\%, exceeding GRPO (82.14\%) by 2.07~pp and SFT (83.93\%) by 0.28~pp.
Edited-side accuracy also increases from 62.64\% with GRPO to 65.93\% with \horn{}.
\backtomain{sec:targeted-scene-edits}

\FloatBarrier
\Needspace{0.82\textheight}
\section{Ego-Visible Results}
\label{app:ego-results}

\begin{table}[H]
\centering
\small
\setlength{\tabcolsep}{8pt}
\caption{Ego-Visible accuracy (\%).}
\label{tab:ego-results}
\begin{tabular}{@{}lrrr@{}}
\toprule
Method & Challenge & Control & Overall \\
\midrule
SceneCOT Zero-shot & 44.58 & 19.64 & 32.11 \\
SceneCOT SFT (20\%) & 50.64 & \textbf{54.34} & 52.49 \\
SceneCOT GRPO & 53.89 & 52.36 & 53.13 \\
\horn{} & \textbf{58.38} & 52.10 & \textbf{55.24} \\
\bottomrule
\end{tabular}
\end{table}

\begin{table}[H]
\centering
\small
\setlength{\tabcolsep}{3pt}
\caption{Full Scene and Ego-Visible coverage by challenge category. Counts and retention rates are per-question means.}
\label{tab:ego-coverage}
\begin{tabular}{@{}lrrrr@{}}
\toprule
Category & \makecell{Points ($10^4$)\\Full Scene $\to$ Ego-Visible} & \makecell{Point\\retention (\%)} & \makecell{Instances\\Full Scene $\to$ Ego-Visible} & \makecell{Key-instance point\\retention (\%)} \\
\midrule
U & $15.65\to6.52$ & 44.83 & $32.25\to25.04$ & 72.55 \\
N & $17.09\to7.01$ & 44.03 & $35.55\to28.11$ & 75.41 \\
I & $15.74\to8.31$ & 56.28 & $31.66\to28.27$ & -- \\
C & $17.38\to6.98$ & 43.36 & $37.38\to28.52$ & 84.90 \\
O & $15.06\to6.73$ & 47.93 & $30.85\to24.96$ & 86.33 \\
R & $16.37\to6.63$ & 44.13 & $33.82\to26.33$ & 85.25 \\
\(\mathrm{N}^{\prime}\) & $18.53\to7.21$ & 41.87 & $41.18\to30.79$ & 73.15 \\
S & $19.34\to7.53$ & 41.64 & $40.77\to29.80$ & 63.88 \\
\bottomrule
\end{tabular}
\end{table}

Table~\ref{tab:ego-results} reports accuracy on the complete Ego-Visible test set: 1,568 Challenge and 1,568 Control questions, totaling 3,136. Post-training results are three-run means.
Structured evidence and response obligations are recomputed after visibility filtering; the original split membership is retained only for aggregate reporting.
Figure~\ref{fig:ego-visible}(b) compares Ego-Visible and Full Scene accuracy.

\textbf{Observation coverage.}
Table~\ref{tab:ego-coverage} compares Full Scene and Ego-Visible coverage over 196 questions per challenge category.
Point counts precede model input sampling, and instance counts include structural annotations such as walls and floors.
Retention ratios are computed per question and then averaged.
Key instances are the designated evidence objects: candidates in U/N, containers in C, anchors in O, attribute targets in R, relation pairs in N$^\prime$, and receptacles/items in S.
I has no in-scan target instance. O's key-instance statistic uses the 150 questions with anchors.
\backtomain{sec:ego-visible}

\clearpage
\section{Outdoor Scene Transfer Evaluation}
\label{app:outdoor-transfer}

\begin{table}[H]
\centering
\caption{Outdoor Veto Accuracy (\%) for SceneCOT from a single evaluation run per method.
Column-best results are bold.}
\label{tab:outdoor-transfer}
\begin{tabular}{lrrr}
\toprule
Method & Challenge & Control & Overall \\
\midrule
ZS & 6.89 & 37.82 & 22.35 \\
SFT & 15.05 & 41.07 & 28.06 \\
GRPO & 16.65 & \textbf{46.62} & 31.63 \\
\horn{} & \textbf{19.39} & 45.47 & \textbf{32.43} \\
\bottomrule
\end{tabular}
\end{table}

\begin{figure}[H]
\centering
\includegraphics[width=0.94\linewidth]{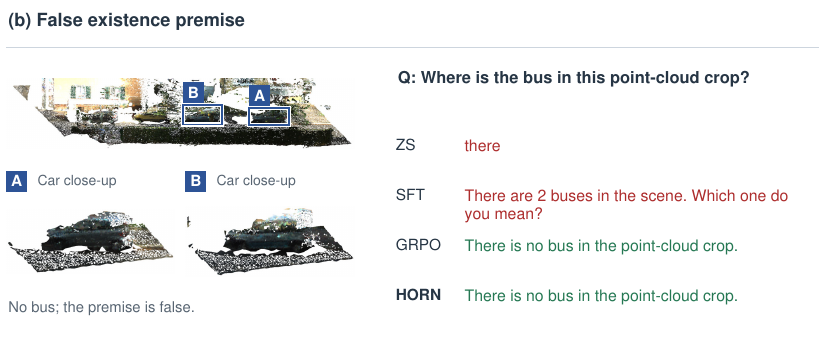}
\caption{Outdoor false-premise example. Green/red responses satisfy/violate
the evidence-veto criteria.}
\label{fig:outdoor-transfer}
\end{figure}

\paragraph{Dataset construction.}
We construct an outdoor evaluation set from the publicly released static
point-cloud annotations of KITTI-360~\citep{Liao2022PAMI}: 652 crops of
$30\,\mathrm{m}\times30\,\mathrm{m}$ from 100 annotated windows across nine
driving sequences. Deduplication uses acquisition sequence and world-coordinate
grid cell, retaining one source window per cell. The 3,136 questions comprise
1,568 Challenge and 1,568 Control, matching the indoor test-set size.
Challenge covers all eight categories: U, N, I, C, O, R, N$^\prime$, and S, with
196 questions per category.

\paragraph{Evidence and evaluation.}
Questions and evidence use instance annotations, point-cloud geometry, and
RGB information. Existence and counting are restricted to the supplied crop;
questions beyond its boundary require acknowledging insufficient observation.
Attribute and relation questions retain construction criteria and supporting
evidence. Model-facing question files exclude reference answers and
target-instance identifiers. We apply the evidence-veto criterion, requiring
both an appropriate response strategy and factual support for all scene
claims. Challenge averages the eight equally sized categories. Control uses
all answerable questions. Overall is their equal-weight mean. Metrics are
computed before rounding and reported to two decimal places.

\paragraph{Results.}
Without outdoor fine-tuning, SFT on \far{} improves outdoor Challenge
accuracy from 6.89\% to 15.05\% and Overall from 22.35\% to 28.06\%
relative to ZS (Table~\ref{tab:outdoor-transfer}). These gains suggest that
\far{} can help models acquire evidence-appropriate response behaviors
that transfer to outdoor scenes. \horn{} further improves Challenge and
Overall to 19.39\% and 32.43\%, respectively, outperforming both ZS and SFT.
Its advantage over GRPO is limited: Challenge increases by 2.74~pp and
Overall by 0.80~pp, while Control decreases by 1.15~pp.

\paragraph{Qualitative example.}
Figure~\ref{fig:outdoor-transfer} shows a question that presupposes a bus
in a crop containing no bus. ZS answers ``there,'' while SFT invents two
buses and asks for clarification. Both GRPO and \horn{} correctly reject
the false premise. Blue A/B boxes identify the cars shown in the close-ups;
model responses are reproduced verbatim.

\FloatBarrier
\Needspace{12\baselineskip}
\section{Full-Scene versus Text-Only Results}\label{app:text-only}

\begin{table}[H]
\centering
\caption{Full-scene versus text-only accuracy (\%).}
\label{tab:text-only}
\small
\setlength{\tabcolsep}{6pt}
\begin{tabular}{lrrrrrr}
\toprule
 & \multicolumn{3}{c}{Full scene} & \multicolumn{3}{c}{Text-only} \\
\cmidrule(lr){2-4}\cmidrule(lr){5-7}
Model & Challenge & Control & Overall & Challenge & Control & Overall \\
\midrule
SceneCOT SFT & 68.6 & 79.2 & 73.9 & 32.65 & 45.28 & 38.97 \\
SceneCOT GRPO & 70.0 & 78.1 & 74.0 & 32.78 & 45.22 & 39.00 \\
\horn{} & \textbf{75.1} & \textbf{79.3} & \textbf{77.2} & \textbf{33.16} & \textbf{45.41} & \textbf{39.29} \\
\bottomrule
\end{tabular}
\end{table}

To examine the dependence of model performance on 3D scene input, we compare full-scene input with a text-only setting that retains only the question text.
Table~\ref{tab:text-only} reports Challenge, Control, and Overall accuracy as three-run means.
Removing scene input reduces Challenge accuracy to approximately 33\% for all three methods and narrows \horn{}'s advantage over GRPO from 5.1~pp with full-scene input to 0.38~pp with text-only input.
These results suggest that \horn{}'s additional gains on Challenge depend on scene evidence, while the nonzero text-only scores indicate that question wording and language priors may also contribute to performance.
\backtomain{sec:text-only}

\end{document}